\RequirePackage{fix-cm}
\documentclass[twocolumn]{svjour3}          % twocolumn
\smartqed  % flush right qed marks, e.g. at end of proof
\usepackage{graphicx}
\usepackage{latexsym}
\usepackage{bm}
\usepackage{times}
\usepackage{epsfig}
\usepackage{graphicx}
\usepackage{amsmath}
\usepackage{amssymb}
\usepackage{subfigure}
\usepackage{multicol}
\usepackage{multirow}
\usepackage{algorithm}
\usepackage{algorithmic}
\usepackage{caption}
\usepackage[flushleft]{threeparttable}
\usepackage{color}
\usepackage{multirow}
\usepackage{array}
\usepackage{hyperref}
\usepackage{amssymb}
\usepackage{nccbbb}
\usepackage{color}

\emergencystretch=\maxdimen
\begin{document}

\title{Open-set Adversarial Defense with Clean-Adversarial Mutual Learning}


\author{Rui Shao         \and
        Pramuditha Perera \and
        Pong C. Yuen \and
        Vishal M. Patel
        %etc.
}


\institute{Rui Shao \at
				Hong Kong Baptist University, Hong Kong, China \\
              ruishao@comp.hkbu.edu.hk           %  \\
           \and
           Pramuditha Perera \at
           	AWS AI Labs, USA		\\
            pramudi@amazon.com
           \and 
           Pong C. Yuen \at
           Hong Kong Baptist University, Hong Kong, China \\
      	   pcyuen@comp.hkbu.edu.hk
           \and 
		   Vishal M. Patel \at
		   Johns Hopkins University, USA \\
		   vpatel36@jhu.edu     
}

\date{Received: date / Accepted: date}

\maketitle

\begin{abstract}
Open-set recognition and adversarial defense study two key aspects of deep learning that are vital for real-world deployment. The objective of open-set recognition is to identify samples from open-set classes during testing, while adversarial defense aims to robustify the network against images perturbed by imperceptible adversarial noise. This paper demonstrates that open-set recognition systems are vulnerable to adversarial samples.  Furthermore, this paper shows that adversarial defense mechanisms trained on known classes are unable to generalize well to open-set samples. Motivated by these observations, we emphasize the necessity of an Open-Set Adversarial Defense (OSAD) mechanism. This paper proposes an Open-Set Defense Network with Clean-Adversarial Mutual Learning (OSDN-CAML) as a solution to the OSAD problem.  The proposed network designs an encoder with dual-attentive feature-denoising layers coupled with a classifier to learn a noise-free latent feature representation, which adaptively removes adversarial noise guided by channel and spatial-wise attentive filters. Several techniques are exploited to learn a noise-free and informative latent feature space with the aim of improving the performance of adversarial defense and open-set recognition. First, we incorporate a decoder to ensure that clean images can be well reconstructed from the obtained latent features. Then, self-supervision is used to ensure that the latent features are informative enough to carry out an auxiliary task.  Finally, to exploit more complementary knowledge from clean image classification to facilitate feature denoising and search for a more generalized local minimum for open-set recognition, we further propose clean-adversarial mutual learning, where a peer network (classifying clean images) is further introduced to mutually learn with the classifier (classifying adversarial images). We propose a testing protocol to evaluate OSAD performance and show the effectiveness of the proposed method on white-box attacks, black-box attacks, as well as the rectangular occlusion attack in multiple object classification datasets.
\end{abstract}

\section{Introduction}
The advent of deep convolutional neural networks (CNNs)~\cite{Kaiming_Resnet_CVPR2016} has contributed to significant improvements in various image classification tasks. Many real-world computer vision applications~\cite{arxiv20reidsurvey,cvpr19uel,Tracking_2019_TIE,Yash2019anomaly,pami20isif,2018TIFSdynamictext,Shao2019CVPR,shao2020regularized} have been realized due to the promising performance of CNNs in classification tasks.  However, there exist several limitations of conventional CNNs that have an impact in real-world applications. In particular, open-set recognition~\cite{bendale2016towards,ge2017generative,neal2018open,oza2019c2ae,yoshihashi2019classification,Perera_2020_CVPR,Perera2019Noveltytransfer,Perera_CVPR19_2,zhang2016sparse} and adversarial attacks~\cite{goodfellow2014explaining,madry2017towards,carlini2017towards,kurakin2016adversarial,xie2019feature} have gained a lot of interest in the computer vision and machine learning communities in the last few years.

\begin{figure*}[!htbp] 
	
	\begin{center}
		
		\includegraphics[height=4.5cm, width=1\linewidth]{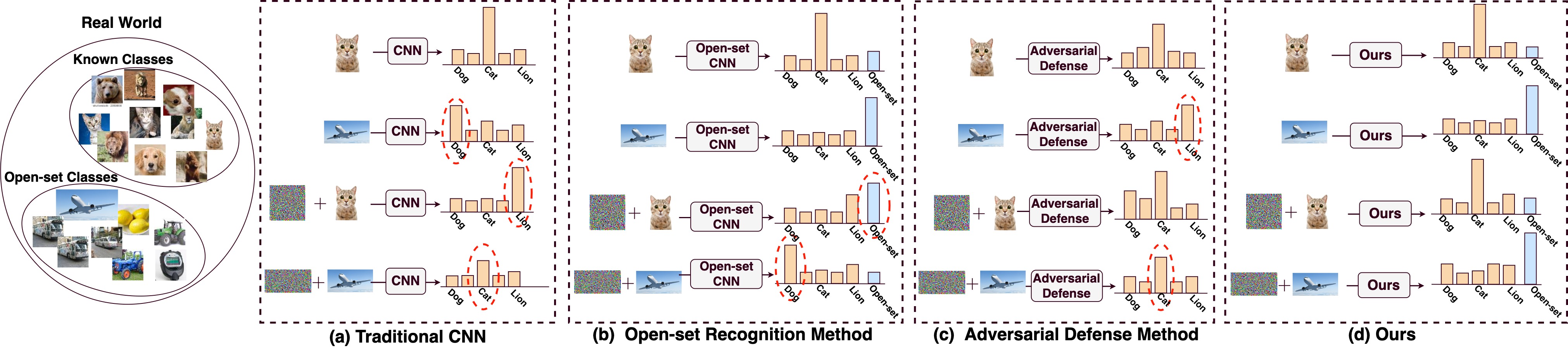}
		
	\end{center}
	
	\vskip-10pt	\caption{Challenges in open-set recognition and adversarial defense. (a) Conventional CNN classifiers perform poorly in the presence of both open-set and adversarial images. (b) Open-set recognition methods are able to identify open-set samples, but cannot generalize well to adversarial samples. (c) Adversarial defense methods fail to identify open-set samples. (d) The proposed method can successfully classify known samples and detect open-set images under adversarial perturbations.}
	\label{fig:overview}
\end{figure*}

\begin{table}[!htbp]
	\centering
	\caption{Importance of an Open-set Adversarial Defense (OSAD) mechanism.}                
	\resizebox{1\linewidth}{!}{
		\normalsize
		\begin{tabular}{|c|c|c|c|l|}
			\hline
			& Clean Images     & \multicolumn{3}{c|}{Adversarial Images}                \\ \hline
			& Original Network & Original Network       & \multicolumn{2}{c|}{Proposed Method} \\ \hline
			Closed Set Accuracy          & 92.79             & 8.65                  & \multicolumn{2}{c|}{74.14}     \\ \hline
			Open-set Detection (AUC-ROC) & 83.72             & 45.98                 & \multicolumn{2}{c|}{73.72}     \\ \hline
		\end{tabular}
	}
	\label{tbl:init}
\end{table}
Conventionally, a CNN assumes that the classes encountered during testing will be identical to those observed during training. But in a real-world scenario, some \textit{open-set} samples from classes unseen during training are likely to be presented to a trained classifier. In this case, the CNN will wrongly classify an open-set sample with a known-set class. Consider a CNN trained with animal classes. Given an input that is from an animal class (such as a cat), the network is capable of predicting the correct classes as shown in Figure~\ref{fig:overview}(a-First Row). However, as illustrated in Figure~\ref{fig:overview}(a-Second Row), when the network encounters a non-animal image, such as an Airplane image, the CNN erroneously classifies it as one of the known classes. On the other hand, it is a well known fact that adding crafted human-imperceptible perturbations to clean images can alter model prediction in a classifier \cite{goodfellow2014explaining}.   These \textit{adversarial attacks} can be easily deployed and threaten various real-world applications~\cite{evtimov2017robust,wu2019defending}.  As shown in Figure~\ref{fig:overview}(a-Third Row) and Figure~\ref{fig:overview}(a-Fourth Row), model predictions for known and open-set images are severely degraded by such adversarial attacks, respectively.

Several open-set recognition algorithms~\cite{bendale2016towards,ge2017generative,neal2018open,oza2019c2ae,yoshihashi2019classification} have been proposed to address the former challenge in the computer vision community. By treating open-set classes as an additional class, these algorithms convert the $c$-class classification problem into a $c+1$ class problem. As shown in Figure~\ref{fig:overview}(b-First and Second rows), correct classification decisions for both known and open-set classes can be provided by these algorithms. However, as illustrated in Figure~\ref{fig:overview}(b-Third and Fourth rows), in the presence of adversarial attacks, these models fail to produce correct predictions. On the other hand, several defense strategies~\cite{kurakin2016adversarial,xie2019feature,liao2018defense,jang2019adversarial} have been developed to combat against the latter challenge. However, these defense mechanisms are all built based on the assumption of closed-set testing. Therefore, although they perform well when this assumption holds (Figure~\ref{fig:overview}(c-First and third rows)), they cannot generalize well to open-set samples as shown in Figure~\ref{fig:overview}(c-Second and Fourth rows).

Based on the above discussion, it is evident that existing open-set recognition algorithms are not robust to adversarial attacks and adversarial defense mechanisms trained on known classes fail to generalize well in the presence of open-set samples. This observation motivates us to propose a new research problem -- Open-Set Adversarial Defense (OSAD), where the objective is to exploit the complementarity between adversarial robustness and open-set generalization such that we can simultaneously detect open-set samples and classify known classes in the presence of adversarial noise. To demonstrate the significance of the proposed problem, a preliminary experiment is conducted in CIFAR10 dataset, where only 6 classes are considered to be known classes to the classifier. We tabulate both open-set detection performance (\textit{i.e.} area under the curve of the ROC curve) and closed-set classification accuracy in Table~\ref{tbl:init} for this experiment. When clean images are presented to the network, a performance better than $80\%$ accuracy in both open-set detection and close-set classification can be achieved. However, when images are attacked by adversarial noises, significant performance drops happen in open-set detection along with the closed set classification. Note that in this case, open-set detection performance is close to random guessing $(45.98\%)$.

To address this new research problem, this paper proposes an Open-Set Defense Network with Clean-Adversarial Mutual Learning (OSDN-CAML) that learns a noise-free, informative latent feature space with the objective of generalizing to open-set samples and being robust to adversarial attacks. We use an autoencoder network with a classifier branch attached to its latent space as the backbone of our solution. Dual-attentive feature denoising layers are embedded into the encoder network to remove adversarial noise guided by spatial and channel-wise attentive filters simultaneously. To improve the informativeness of the learned feature space, clean image generation and self-supervised denoising are incorporated into our network, which facilitate the detection of open-set samples under adversarial attacks. Clean image generation generates noise-free images based on the learned latent features through a decoder, and self-supervised denoising is carried out by forcing the network to perform an auxiliary classification task based on the learned noise-free features. Moreover, to fully exploit the complementarity between clean images and their corresponding adversarial examples to aid adversarial defense and open-set recognition, we incorporate a peer learner (classifying clean images) to mutually learn with the classifier (classifying adversarial images). The proposed clean-adversarial mutual learning can further exploit more complementary knowledge from clean images classification to facilitate feature denoising and search for a more generalized local minimum for open-set recognition. As shown in Table~\ref{tbl:init}, the proposed OSDN-CAML significantly improves the robustness against adversarial attacks in terms of close-set classification as well as open-set detection. Main contributions of our paper are summarized as follows:

\noindent 1. This paper proposes a new research problem named Open-Set Adversarial Defense (OSAD) where adversarial attacks are studied under an open-set setting.

\noindent 2. We propose an Open-Set Defense Network with Clean-Adversarial Mutual Learning (OSDN-CAML) that learns a latent feature space being robust to adversarial attacks and informative to identify open-set samples. Dual-attentive denoising layers are embedded in encoder for better feature denoising and clean-adversarial mutual learning is proposed to exploit the complementarity between clean images and their corresponding adversarial examples to aid adversarial defense and open-set recognition.

\noindent3. A test protocol is defined to the OSAD problem. Extensive quantitative experiments including white-box attacks, black-box attacks, and rectangular occlusion attacks are conducted on three publicly available image classification datasets to demonstrate the effectiveness of the proposed method. Various qualitative visualizations provide more comprehensive analysis and understanding of the proposed method.

A preliminary version of this work appeared in ECCV 2020~\cite{Shao_2020_OSAD}.  We have made three major improvements in this journal version. Firstly, we propose a dual-attentive feature denoising layer to improve feature denoising in the encoder. Compared to denoising operation carried out in our conference version which focuses on removing adversarial noise only in the spatial dimension, the proposed dual-attentive feature denoising layers carry out both channel and spatial-wise feature denoising guided by channel and spatial-wise attentive filters. When dual-attentive denoising layers are used, the encoder can simultaneously learn 'where' and 'what' to emphasize or suppress for feature denoising. Secondly, we propose  clean-adversarial mutual learning in which  we further incorporate one more branch, peer learner for clean image classification, into our framework. In the proposed clean-adversarial mutual learning, the peer learner and encoder-open-set classifier branch mutually learn with each other, which facilitates feature denoising and helps the network to converge to a minimum with better generalization ability to open-set samples. Finally, more comprehensive quantitative and qualitative experiments are carried out in this journal version for white-box attacks, black-box attacks, as well as the rectangular occlusion attack to demonstrate the effectiveness of the proposed method.

\section{Related Work}

\noindent \textbf{Adversarial Attack and Defense Methods.}
Szegedy \textit{et al.}~\cite{szegedy2013intriguing} demonstrate that by adding crafted human-imperceptible perturbations, adversarial attacks can mislead CNNs into making incorrect predictions. Fast Gradient Sign Method (FGSM)~\cite{goodfellow2014explaining} is proposed to generate adversarial samples by calculating the sign of a gradient update from the classifier. Basic Iteration Method (BIM)~\cite{kurakin2016adversarial} and Projected Gradient Descent (PGD)~\cite{madry2017towards} extend FGSM into iterative versions to form stronger attacks. Different from the above gradient-based adversarial attacks, Carlini and Wagner~\cite{carlini2017towards} propose the C$\&$W attack to take a direct optimization approach to generate adversarial samples. Among various adversarial defense mechanisms, adversarial training~\cite{madry2017towards}, as one of the most popular defense methods, improves the robustness of the network by training it with adversarial images generated on-the-fly based on network's current parameters. Lately, several adversarial defense methods have been developed to further improve adversarial training with denoising-based operations. Pixel denoising~\cite{liao2018defense} is proposed to guide the denoising process using the high-level features. The method proposed in~\cite{gupta2019ciidefence} carries out  pixel-level denoising by exploring the most influential local parts based on class activation map responses. Xie \textit{et al.} ~\cite{xie2019feature} exploit  adversarial noise removal in the feature space using denoising filters.

\noindent \textbf{Open-set Recognition.} 
The possibility for open-set samples to generate very high probability scores in a closed-set classifier is first brought to attention in \cite{Scheirer_2013_TPAMI}. Deep learning models are also shown to be affected by the same phenomena in ~\cite{bendale2016towards}. The method, called  OpenMax~\cite{bendale2016towards}, is proposed with a statistical solution for this problem, where the normal $c$-class classification problem is converted into a $c+1$ problem by considering the extra class to be the open-set class. The logits of known classes to the open-set class are apportioned by considering spatial positioning of a query sample in an intermediate feature space. This idea is further extended by~\cite{ge2017generative} and \cite{neal2018open} to exploit a generative model to produce logits of the open-set class. Yoshihashi \textit{et al.}~\cite{yoshihashi2019classification} argue that a generative feature contains more information facilitating open-set recognition. On these grounds, a concatenation of a generative feature and a classifier feature is considered when designing the OpenMax layer. Authors in~\cite{oza2019c2ae} exploit a generative approach to design a class conditioned decoder for open-set detection. Works of both~\cite{oza2019c2ae} and~\cite{yoshihashi2019classification} demonstrate that the open-set recognition can be benefited from generative features. It should be noted that open-set recognition is more challenging than novelty detection~\cite{Poojan2020NoveltyDistribution,Poojan2020NoveltyPatch,Perera2019Noveltytransfer,perera2019learning,oza2018one} because novelty detection only requires determining whether an observed image during inference belongs to one of the known classes.

\noindent \textbf{Self-Supervision.}  
Self-supervision is an unsupervised machine learning technique where the data itself is used to provide supervision. Several techniques have been introduced to facilitate performance in classification and detection tasks. For example, given an anchor image patch, the method proposed in ~\cite{selfsup} carries out self-supervision by requiring the network to predict the relative position of a second image patch. Authors in~\cite{Doersch_2017_ICCV} extend this idea with a multi-task prediction framework, in which the network is forced to predict a combination of relative order and pixel color. In the work~\cite{gidaris2018unsupervised}, the network is trained to predict the angle of the transformed images which are randomly rotated by a factor of 90 degrees.

\section{Background}
\begin{figure*}[t] 
	\begin{center}	
		\includegraphics[ width=.9\linewidth]{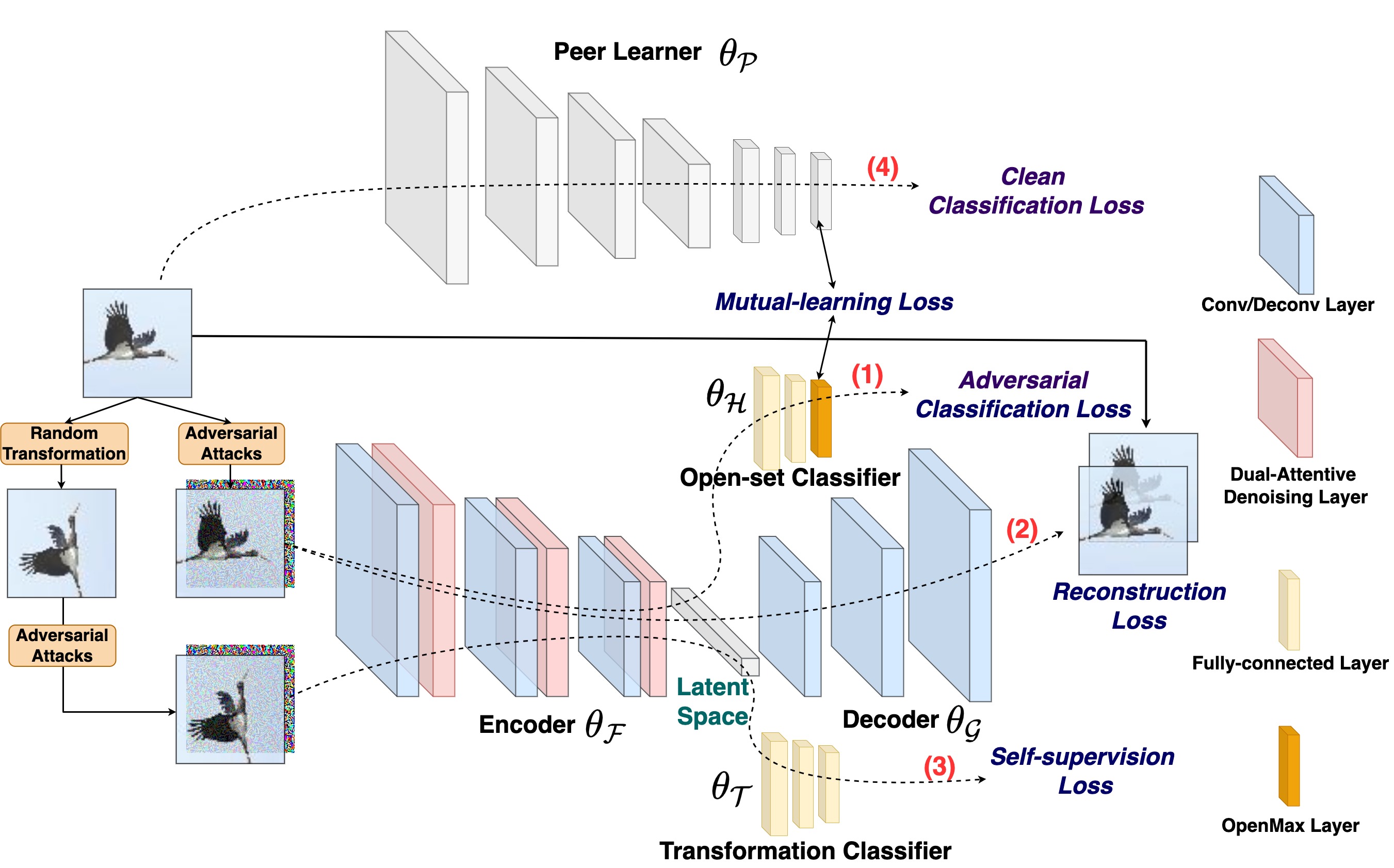}	
	\end{center}
	\vskip -10pt	\caption{{Network structure of the proposed Open-Set Defense Network with Clean-Adversarial Mutual Learning (OSDN-CAML). It consists of five components: encoder, decoder, open-set classifier, transformation classifier, and peer learner.}}
	\label{fig:nt}
\end{figure*}

\noindent \textbf{Adversarial Attacks.} Consider a network parameterized by parameters $\theta$. Provided by data and label pairs $(\textbf{x}, \textbf{y})$, we can generate adversarial images $\textbf{x}^{adv}$ via $\textbf{x}^{adv} = \textbf{x} + \delta$, where $\delta$ can be determined by a given white-box attack based on the models parameters. In this paper, two types of white-box adversarial attacks are considered.

The first attack considered is the Fast Gradient Signed Method (FGSM)~\cite{goodfellow2014explaining} where the adversarial images are formed as follows,
\setlength{\belowdisplayskip}{0pt} \setlength{\belowdisplayshortskip}{0pt}
\setlength{\abovedisplayskip}{0pt} \setlength{\abovedisplayshortskip}{0pt}
\begin{equation}
	\textbf{x}^{adv} = {\rm Proj}_\chi (\textbf{x}+\epsilon sign(\bigtriangledown_\textbf{x} \mathcal{L}(\textbf{x}, \textbf{y}; \theta))),
\end{equation} 
where $\mathcal{L}(\cdot)$ is a classification loss. ${\rm Proj}_\chi$ denotes the projection of its element to a valid pixel value range, and $\epsilon$ denotes the size of $l_\infty$-ball. The second attack considered is Projective Gradient Descent (PGD) attacks~\cite{madry2017towards}. Adversarial images are generated in this method as follows,
\begin{equation}
	\textbf{x}^{adv(t+1)} = {\rm Proj}_{\zeta\cap\chi} (\textbf{x}^{adv(t)}+\epsilon_{step} sign(\bigtriangledown_\textbf{x} \mathcal{L}(\textbf{x}^{adv(t)}, \textbf{y}; \theta))),
\end{equation} 
where ${\rm Proj}_{\zeta\cap\chi}(\cdot)$ means the projection of its element to $l_\infty$-ball $\zeta$ and a valid pixel value range, and $\epsilon_{step}$ represents a step size smaller than $\epsilon$. The adversarial samples of the final step $T$: $\textbf{x}^{adv} = \textbf{x}^{adv(T)}$ are used in this work.\\

\noindent \textbf{OpenMax Classifier.} Typically, $c$ probability predictions correspond to a SoftMax classifier trained for a $c$-class problem. OpenMax extends them into the probability scores of $c+1$ classes, where the probability of the final class represents the open-set class. Given $c$ known classes $\mathcal{K}=\{C_1, C_2, ..., C_c \}$, to identify open-set samples, OpenMax is designed to calibrate the final hidden layer of a classifier as follows:
\begin{equation}
	\hat{\textbf{\textit{l}}}_i=
	\begin{cases}
		\textbf{\textit{l}}_i \textbf{\textit{w}}_i& (i\leq c)\\
		\textbf{$\sum^{c}_{i=1}$ \textit{l}}_i (1-\textbf{\textit{w}}_i)& (i= c+1),
	\end{cases}
\end{equation}
\begin{equation}
	{\rm OpenMax}_i(\textbf{x})={\rm SoftMax}_i(\hat{\textbf{\textit{l}}}),
\end{equation}
where $\textbf{\textit{l}}$ denotes the logit vector obtained prior to the SoftMax operation in  a classifier, the belief that $x$ belongs to the known class $C_i$ is represented by $\textbf{\textit{w}}_i$. Here, open-set class corresponds to the class $C_{c+1}$. Belief $\textbf{\textit{w}}_i$ is quantified by calculating the distance between a given sample and it's class mean $\mu$ in an intermediate feature space. During training, a matched score distribution is formed by calculating the distance of all training image samples from a given class to its corresponding class mean $\mu$. Then, a Weibull distribution is employed to fit the tail of the matched distribution. Assuming the feature representation of the input in the same feature space is $\textbf{\textit{v(x)}}$, $\textbf{\textit{w}}_i$ can be calculated as follows,
\begin{equation}
	\begin{split}
		\textbf{\textit{w}}_i = 1 - \max\Big(0, \frac{\sigma-{\rm rank}(i)}{\sigma}\Big)e^{\Big( -\Big(\dfrac{|\textbf{\textit{v(x)}}-\mu_i|_2}{\eta_i}\Big)^{m_i}\Big)},
	\end{split}
\end{equation}
where $m_i, \eta_i$ are parameters of the Weibull distribution that corresponding to class $C_i$. $\sigma$ is hyperparameter and ${\rm rank}(i)$ is the index in the logits sorted in the descending order.

\section{Proposed Method}

The proposed network consists of five components: encoder, decoder, open-set classifier, transformation classifier and peer learner. Figure~\ref{fig:nt} shows the network structure of the proposed method and illustrates computation flow. The encoder network is embedded with several dual-attentive denoising layers between the convolutional layers. Open-set classifier is structurally similar to a regular classifier, but an OpenMax layer is added on the top of the classifier during inference, which is denoted by an OpenMax layer in Figure~\ref{fig:nt}. A normal CNN is further incorporated as the peer learner. We include the encoder, decoder, open-set classifier, transformation classifier as the main part of our proposed network.

An adversarial image is firstly generated based on the corresponding input clean image. The latent feature of this image is obtained by passing it through the encoder network. An adversarial classification loss $\mathcal{L}_{cls(adv)}$ is calculated by passing this feature through the open-set classifier via path (1). Then, we pass the feature through the decoder via path (2) to generate the corresponding image, which is used to measure its distance to the corresponding clean image to calculate the reconstruction loss $\mathcal{L}_{rec}$. Moreover, we employ a geometric transform on the input image and generate a corresponding adversarial transformed image. This image goes through path (3) to arrive at the transformation classifier and performs self-supervision loss $\mathcal{L}_{ssd}$ by considering the transform applied to the image. Finally, the clean image is passed through peer learner via path (4) to evaluate the clean classification loss $\mathcal{L}_{cls(clean)}$. The clean-adversarial mutual learning is carried out using the Kullback Leibler Divergence based mutual-learning loss $\mathcal{L}_{mut}$ to match the probability predictions between open-set classifier and peer learner. The whole optimization process is iteratively carried out between the main part and the peer learner of the proposed network using the following loss functions:

\begin{equation}
	\mathcal{L}_{OSDN-CAML(Main)} = \mathcal{L}_{cls(adv)} + \mathcal{L}_{rec} + \mathcal{L}_{ssd} + \mathcal{L}_{mut}
\end{equation} 
and
\begin{equation}
	\mathcal{L}_{OSDN-CAML(Peer)} = \mathcal{L}_{cls(clean)} + \mathcal{L}_{mut}.
\end{equation}

In the following subsections, we describe various components and computations involved in all four paths in detail.\\

\begin{figure}[t]
	\begin{center}	
		\includegraphics[ width=0.9\linewidth]{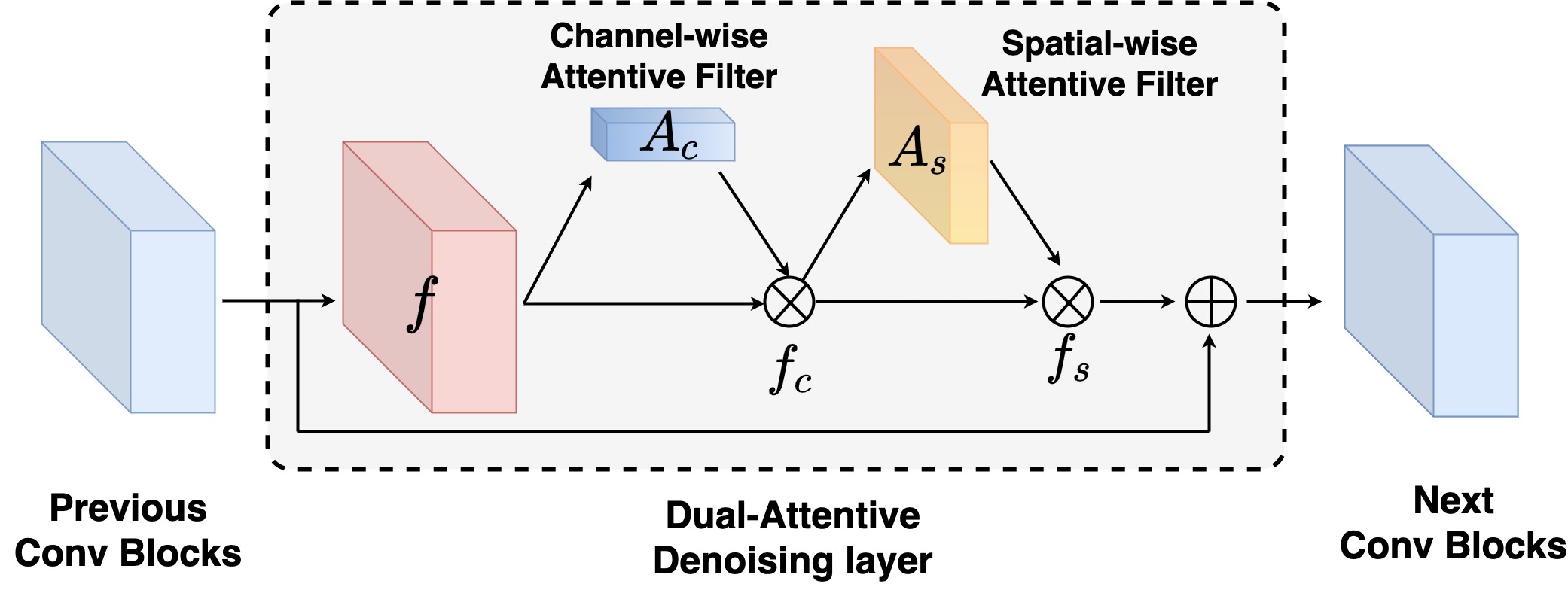}	
	\end{center}
	\vskip-10pt	\caption{{Dual-Attentive Denoising Layer.}}
	\label{fig:DDL}
\end{figure}

\noindent \textbf{Dual-Attentive Noise-free Feature Encoding.} The proposed network learns the noise-free features via an encoder network. Then, the open-set classifier is utilized to perform classification based on the learned features. Inspired by~\cite{xie2019feature}, feature denoising layers are embedded after each main convolutional blocks in the encoder so that feature denoising can be explicitly carried out on adversarial samples. However, we find that variants of feature denoising layers exploited in~\cite{xie2019feature,Shao_2020_OSAD}, such as non-local means~\cite{buades2005non}, bilateral filter~\cite{tomasi1998bilateral}, mean filter and median filter, all focus on removing adversarial noise only in the spatial dimension. The effect of adversarial noise differs between channels and thus we should further emphasize the feature denoising on specific channels which are most heavily contaminated by adversarial noise. Moreover, since spatial and channel information of features complement each other, it is critical to correlate the feature denoising from channel and spatial dimensions and thus exploit an optimal feature denoising operation. Inspired by~\cite{woo2018cbam}, as illustrated in Fig~\ref{fig:DDL}, we propose dual-attentive feature denoising layers that can carry out feature denoising via channel and spatial-wise attentive filters.

It is well known that each channel of a feature map can be regarded as a feature detector~\cite{zeiler2014visualizing} to tell 'what' features are extracted. Given an input adversarial feature map $f\in \mathbb{R}^{C \times H \times W}$, we first calculate a 1D channel-wise attentive filter $A_{c}\in \mathbb{R}^{C \times 1 \times 1}$ to learn what kind of adversarial features should be emphasized for denoising. Considering the efficient calculation of channel attention, average-pooling and max-pooling are adopted to squeeze the spatial dimension of the input feature map, which generates two different spatial context descriptors. Both descriptors are then passed through a shared multi-layer perceptron (MLP) to produce the channel-wise attentive filter as follows:
\begin{equation}
	A_{c}(f)=\sigma({\rm \textbf{MLP}}(AvgPool(f))+{\rm \textbf{MLP}}(MaxPool(f))), 
\end{equation} 
where $\sigma$ denotes the sigmoid function. We can obtain the channel-wise denoised feature map $f_c$ as: $f_c=A_{c}(f) \otimes f$, where $\otimes$ denotes element-wise multiplication. Different from the channel information, the spatial dimension of feature maps encode ‘where’ the informative regions are.  Thus we can calculate a 2D spatial-wise attentive filter $A_{s}\in \mathbb{R}^{H \times W}$ to determine the most critical spatial regions of adversarial features for denoising. Similarly, given channel-wise denoised feature map $f_c$, we first apply average-pooling and max-pooling operations along its channel axis and concatenate them to generate a feature descriptor. Then, we feed the feature descriptor into a convolutional layer (Conv) to generate a spatial-wise attentive filter as follows:
\begin{equation}
	A_{s}(f_c)=\sigma({\rm \textbf{Conv}}([AvgPool(f_c);MaxPool(f_c)])).
\end{equation} 
We obtain the final denoised feature map $f_s$ as: $f_s=A_{s}(f_c) \otimes f_c = A_{s}(A_{c}(f) \otimes f) \otimes(A_{c}(f) \otimes f) $, where it can be seen that the input adversarial features are denoised from both spatial and channel dimensions simultaneously.

Through these dual-attentive denoising layers, the encoder is able to exploit complementary attention from channel and spatial dimensions to simultaneously learn 'where' and 'what' to emphasize or suppress for feature denoising. Formally, we denote the encoder embedded with dual-attentive denoising layers as $\mathcal{F}$ parameterized by $\theta_{\mathcal{F}}$, and the open-set classifier as $\mathcal{H}$ parameterized by $\theta_{\mathcal{H}}$. Given $N$ labeled clean data $\textbf{x}=\{ x_i\}_{i=1}^N $ with $C$ known classes. The corresponding labels are denoted as $\textbf{y}=\{ y_i\}_{i=1}^N $ with $y_i \in \{1, 2,..., C\} $. We can generate the adversarial images $\textbf{x}^{adv}=\{ x_i^{adv}\}_{i=1}^N $ on-the-fly using either FGSM or PGD attacks based on the current parameters $\theta_{\mathcal{F}}$, $\theta_{\mathcal{H}}$ using the true label $\textbf{y}$. The obtained adversarial images $\textbf{x}_{adv} $ are then passed through encoder and open-set classifier (via path (1)) and the probability of class $k$ for $ x_i^{adv}$ given by the encoder-open-set classifier branch can be computed as:
\begin{equation}
	p^k(x_i^{adv}; \theta_{\mathcal{F}}, \theta_{\mathcal{H}}) = \frac{{\rm exp}(z_{\mathcal{F},\mathcal{H}}^k)}{\sum\limits_{k=1}^{C}{\rm exp}(z_{\mathcal{F},\mathcal{H}}^k)},
\end{equation} 
where $z_{\mathcal{F},\mathcal{H}}^k$ is the output logit of the concatenated branch of encoder and open-set classifier. With the calculated probability, the cross-entropy loss based adversarial training of this branch can be defined as:
\begin{equation}
	\begin{split}
		&\mathcal{L}_{cls(adv)} =\mathcal{L}_{CE}(\textbf{x}_{adv}, \textbf{y}; \theta_{\mathcal{F}}, \theta_{\mathcal{H}})\\
		&=-\sum\limits_{i=1}^{N}\sum\limits_{k=1}^{C}\bbbone[k=y_i]logp^k(x_i^{adv}; \theta_{\mathcal{F}}, \theta_{\mathcal{H}}).
	\end{split}
\end{equation}

By minimizing the above adversarial classification loss, a noise-free latent feature space can be learned by the trained encoder embedded with the dual-attentive denoising layers. An OpenMax layer is integrated on top of the classifier to do inference. In this case, even when the input is contaminated with adversarial noise, the open-set classifier operating on the noise-free latent feature is still capable of predicting the correct classes for known samples and simultaneously detecting open-set samples.\\

\noindent \textbf{Clean Image Generation.}
The image generation branch targets at generating noise-free images from adversarial images by integrating the decoder network. This is motivated by two rationales. First, the structure of autoencoders is widely exploited in the literature for image denoising. By training the autoencoder network for noise-free images generation, additional supervision can be provided for noise removal in the latent feature space. Secondly, it is well known that more descriptive features facilitate open-set recognition~\cite{yoshihashi2019classification}. A classifier only learns to model the boundary of each class when trained with class labels. In this case, a feature produced by a classification network only contains information that is necessary for class boundaries modeling. However, when the network is further trained to generate noise-free images based on the latent representations, it ends up with learning generative features. As a result, features become more descriptive than in the case of a pure classifier. In fact, existing open-set recognition works~\cite{yoshihashi2019classification} and~\cite{oza2019c2ae} also exploit such generative features to improve the performance of open-set recognition. Therefore, we argue that incorporating an image generation branch with a decoder can benefit both open-set recognition and adversarial defense.

Based on the above idea, as shown in Figure~\ref{fig:nt}, we pass adversarial images through path (2) to decode the images from latent features. The decoder network denoted as $\mathcal{G}$ parameterized by $\theta_{\mathcal{G}}$ and the encoder network $\mathcal{F}$ are jointly optimized to  minimize the distance between the generated images and the corresponding clean images using the following mean-square error-based reconstruction loss:
\begin{equation}
	\mathcal{L}_{rec} = \mathcal{L}_{MSE}(\textbf{x}, \textbf{x}_{adv}; \theta_{\mathcal{F}}, \theta_{\mathcal{G}}) = \sum\limits_{i=1}^{N} \|x_i-{\mathcal{G}}({\mathcal{F}}(x_i^{adv}) \|_2^2.
\end{equation}

\noindent \textbf{Self-supervised Denoising.}
Furthermore, we propose to exploit self-supervision as a means to further increase the robustness and informativeness of the latent feature space. Self-supervision is an unsupervised machine learning technique that learns representations from data itself. Our work adopts rotation-based self-supervision task~\cite{gidaris2018unsupervised}. Specifically, the task proposed in~\cite{gidaris2018unsupervised} applies a random rotation from a finite set of possible rotations to an image. Based on this, a classifier is trained to automatically recognize the applied image rotation. 

In our approach, following \cite{gidaris2018unsupervised}, we first generate a random number $\textbf{r} = \{r_i \}_{i=1}^N \in \{0, 1, 2, 3\}$ as the rotation ground-truth and transform the input clean image $\textbf{x}$ by rotating them with $ 90^\circ r$ degrees denoted as $R_\textbf{r}(\textbf{x})=\{ R_{r_i}(x_i)\}_{i=1}^N$, where $R_{r_i}$ is a rotation transformation. We denote the transformation classifier as $\mathcal{T}$ parameterized by $\theta_{\mathcal{T}}$. Based on the rotated clean image, we generate the rotated adversarial image $R_\textbf{r}(\textbf{x})^{adv}=\{ R_{r_i}(x_i)^{adv}\}_{i=1}^N$ on-the-fly using either FGSM or PGD attack based on the current network parameters $\theta_{\mathcal{F}}$, $\theta_{\mathcal{T}}$ using rotation ground-truth $\textbf{r}$. Obtained adversarial rotated image $R_\textbf{r}(\textbf{x})^{adv}$ is passed through encoder and transformation classifier (via path (3)) and the probability of rotation $k$ for $ R_{r_i}(x_i)^{adv}$ given by the encoder and transformation classifier can be computed as:
\begin{equation}
	p^k( R_{r_i}(x_i)^{adv}; \theta_{\mathcal{F}}, \theta_{\mathcal{T}}) = \frac{{\rm exp}(z_{\mathcal{F},\mathcal{T}}^k)}{\sum\limits_{k=0}^{3}{\rm exp}(z_{\mathcal{F},\mathcal{T}}^k)},
\end{equation} 
where $z_{\mathcal{F},\mathcal{T}}^k$ is the output logit of the concatenated branch of encoder and transformation classifier. Thus we can formulate the adversarial training loss function for self-supervised denoising as follows:
\begin{equation}
	\begin{split}
		&\mathcal{L}_{ssd} = \mathcal{L}_{CE}(R_\textbf{r}(\textbf{x})^{adv}, \textbf{r}; \theta_{\mathcal{F}}, \theta_{\mathcal{T}})\\
		&=-\sum\limits_{i=1}^{N}\sum\limits_{k=0}^{3}\bbbone[k=r_i]logp^k( R_{r_i}(x_i)^{adv}; \theta_{\mathcal{F}}, \theta_{\mathcal{T}}).
	\end{split}
\end{equation}

Exploiting self-supervision in our method is motivated by multiple reasons. Training a classifier to recognize different rotations deepens its understanding on object structures and orientations of known classes. As such, carrying out self-supervision in addition to classification enables the underlying feature space to represent additional information that was not considered in the case of a pure classifier. Therefore, self-supervision enhances the informativeness of the latent feature space which will facilitate the open-set recognition. On the other hand, since we use adversarial images for self-supervision, additional feature denoising based on the transformed adversarial images is carried out, which further contributes towards learning the denoising operator in the encoder. Please note that self-supervised learning facilitating the robustness against adversarial samples has also been found in recent work~\cite{hendrycks2019using}. Based on these factors, we believe that open-set detection and adversarial defense processes are both benefited from the addition of self-supervision.\\

\begin{figure}[t] 
	\begin{center}	
		\includegraphics[ width=1\linewidth]{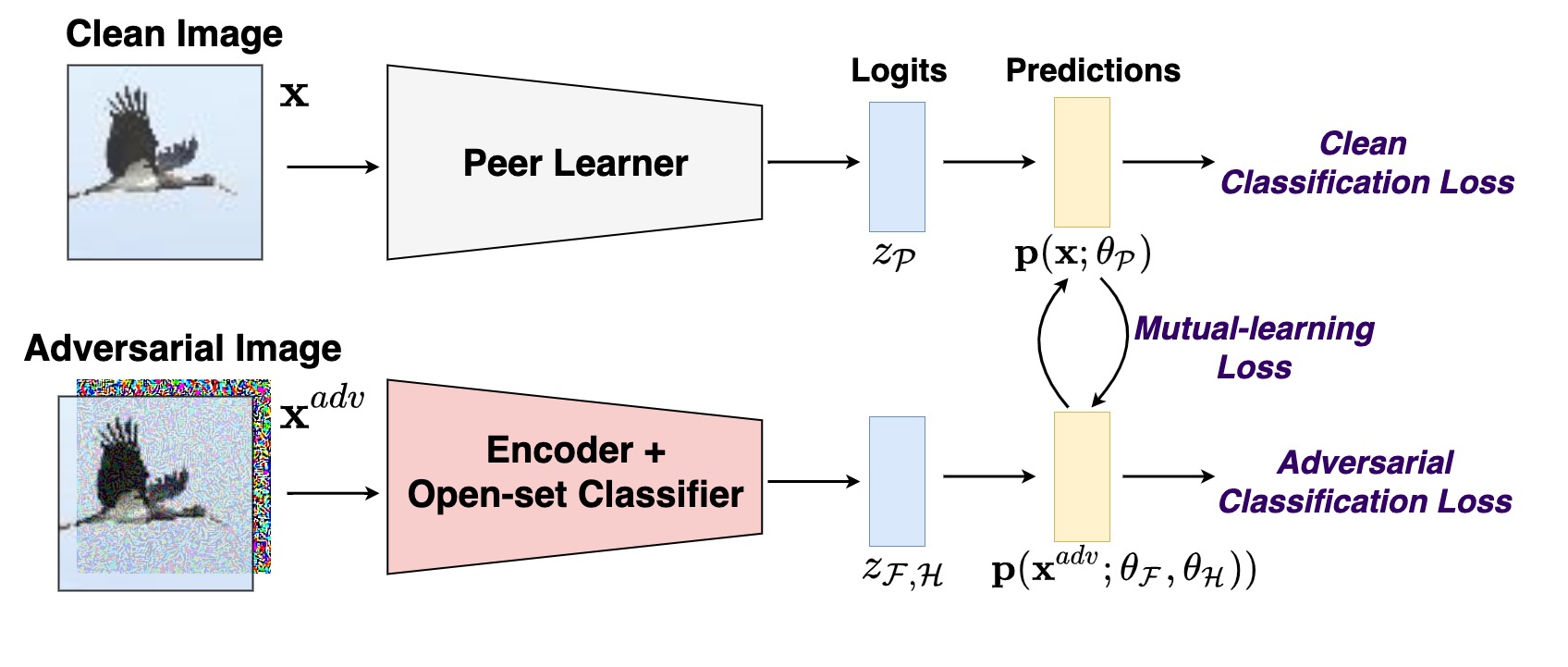}	
	\end{center}
	\vskip-15pt	\caption{{Clean-Adversarial Mutual Learning.}}
	\label{fig:mutual}
\end{figure}

\noindent \textbf{Clean-Adversarial Mutual Learning.} To further facilitate adversarial defense and improve the generalization ability for open-set recognition, we incorporate a peer learner into our framework which learns to classify clean images with supervised training. In our approach, we have an encoder-open-set classifier branch which learns to classify the corresponding adversarial images with adversarial training. As illustrated in Fig~\ref{fig:mutual}, we propose a clean-adversarial mutual learning to let these two branches mutually learn with each other. This is motivated by the following two  reasons. 

First, since the above introduced parts are all based on the inputs of adversarial images for adversarial training, useful information corresponding to clean images is underutilized and worth exploring for adversarial defense. More importantly, directly combining clean and adversarial images to carry out adversarial training has little effect on boosting model robustness~\cite{xie2020intriguing,xie2020adversarial}. Inspired by deep mutual learning~\cite{zhang2018deep}, instead of using single branch mixing up with clean and adversarial images to conduct adversarial training, we introduce a peer learner as an auxiliary branch for clean image classification, and let it mutually learn with the encoder-open-set classifier branch which carries out adversarial training for the corresponding adversarial examples. In the process of this clean-adversarial mutual learning, the encoder-open-set classifier branch and peer learner learn to mimic the predictions with each other and thus they can collaboratively learn to correctly classify the images. In this way, more complementary knowledge corresponding to clean image classification can be exploited from the peer learner to aid the adversarial defense branch to denoise and classify corresponding adversarial images. Second, since the two branches learn to classify the clean and corresponding adversarial version of the same images, their classification objectives are the same. However, they have different network structures with different capabilities and start from different initialization. In this case, they reach the local minima with different gradient descent paths and thus they can share different perspectives corresponding to the same classification objective with each other. Therefore, through knowledge communication by mutual learning with peer learner, the encoder-open-set classifier branch is less likely to be overfitted and its learning process is guided to converge to a more appropriate minimum with better generalization ability to open-set samples. Therefore, we argue that by fully exploiting the complementarity between clean images and their corresponding adversarial examples, the proposed clean-adversarial mutual learning can further benefit both open-set recognition and adversarial defense. 

We denote the peer learner as $\mathcal{P}$ parameterized by $\theta_{\mathcal{P}}$. The probability of class $k$ for the clean image $ x_i$ given by the peer learner can be computed as:
\begin{equation}
	p^k(x_i; \theta_{\mathcal{P}}) = \frac{{\rm exp}(z_{\mathcal{P}}^k)}{\sum\limits_{k=1}^{C}{\rm exp}(z_{\mathcal{P}}^k)},
\end{equation} 
where $z_{\mathcal{P}}^k$ is the output logit of the peer learner. The mutual learning is carried out by matching the probability predictions between the encoder-open-set classifier branch and the peer learner. We use the Kullback Leibler (KL) Divergence as the metric to quantify the probability match and thus we obtain the KL Divergence based mutual-learning loss for the encoder-open-set classifier branch as follows:
%\begin{equation}
%\begin{split}
%&\mathcal{L}_{mut} = D_{KL}(p^k(x_i; \theta_{\mathcal{P}})\|p^k(x_i^{adv}; \theta_{\mathcal{F}}, \theta_{\mathcal{H}}))\\
%&=\sum\limits_{i=1}^{N}\sum\limits_{k=1}^{C}p^k(x_i; \theta_{\mathcal{P}}){\rm log}\frac{p^k(x_i; \theta_{\mathcal{P}})}{p^k(x_i^{adv}; \theta_{\mathcal{F}}, \theta_{\mathcal{H}})}
%\end{split} 
%\end{equation} 

\begin{equation}
	\begin{split}
		&\mathcal{L}_{mut} = D_{KL}(\textbf{p}(\textbf{x}; \theta_{\mathcal{P}})\|\textbf{p}(\textbf{x}^{adv}; \theta_{\mathcal{F}}, \theta_{\mathcal{H}}))\\
		&=\sum\limits_{i=1}^{N}\sum\limits_{k=1}^{C}p^k(x_i; \theta_{\mathcal{P}}){\rm log}\frac{p^k(x_i; \theta_{\mathcal{P}})}{p^k(x_i^{adv}; \theta_{\mathcal{F}}, \theta_{\mathcal{H}})}.
	\end{split} 
\end{equation} 
The above mutual-learning loss can integrated into the optimization of the main part (including encoder, decoder, open-set classifier, transformation classifier) of the proposed network. Similarly, the peer learner learns both to correctly predict labels of clean images and to match the probability predictions as follows:
\begin{equation}
	\begin{split}
		&\mathcal{L}_{OSDN-CAML(Peer)} = \mathcal{L}_{cls(clean)} + \mathcal{L}_{mut}\\
		&=\mathcal{L}_{CE}(\textbf{x}, \textbf{y}; \theta_{\mathcal{P}})+D_{KL}(\textbf{p}(\textbf{x}^{adv}; \theta_{\mathcal{F}}, \theta_{\mathcal{H}})\|\textbf{p}(\textbf{x}; \theta_{\mathcal{P}})).
	\end{split} 
\end{equation}

\section{Experimental Results}
To evaluate the effectiveness of the proposed method, experiments are carried out on four popular multiple-class classification datasets. In this section, we first introduce datasets, baseline methods and describe the protocol used in our experiments. We examine our method and baselines in the task of open-set recognition under adversarial white-box attacks, black-box attacks and rectangular occlusion attacks. In the experiments of white-box attacks, additional experiments regarding the task of out-of-distribution detection are further carried out to demonstrate the effectiveness of our method. We also conduct a detailed ablation study. This section is concluded with various visualizations along with comprehensive analyses.

\begin{table}[!htb]
	\centering
	\small
	\caption{Dataset splits used in SVHN dataset.}
	
	\begin{tabular}{c|p{4cm}<{\centering}}
		\hline
		\multirow{2}{*}{Splits} & \textbf{SVHN}    \\ \cline{2-2} 
		& Known Classes    \\ \hline
		First                   & 0, 1, 2, 4, 5, 9 \\ \hline
		Second                  & 0, 3, 5, 7, 8, 9 \\ \hline
		Third                   & 0, 1, 5, 6, 7, 8 \\ \hline
	\end{tabular}
	\label{tab:svhn}
\end{table}

\begin{table}[!htb]
	\centering
	\small
	\caption{Dataset splits used in CIFAR10 dataset.}
	\begin{tabular}{c|c}
		\hline
		\multirow{2}{*}{Splits} & \textbf{CIFAR10}                             \\ \cline{2-2} 
		& Known Classes                                \\ \hline
		First                   & airplane, automobile, bird, deer, dog, truck \\ \hline
		Second                  & airplane, cat, dog, horse, ship, truck       \\ \hline
		Third                   & airplane, automobile, dog, frog, horse, ship \\ \hline
	\end{tabular}
	\label{tab:cifar10}
\end{table}

\begin{table}[!htb]
	\centering
	\caption{Dataset splits used in TinyImageNet dataset.}
	\begin{tabular}{c|l}
		\hline
		\multirow{2}{*}{Splits} & \multicolumn{1}{c}{\textbf{TinyImageNet}}                                                                                         \\ \cline{2-2} 
		& \multicolumn{1}{c}{Known Classes}                                                                                                 \\ \hline
		First                   & \begin{tabular}[c]{@{}l@{}}143, 94, 155, 109, 27, 102, 131, 43, 194, 186,\\ 56, 24, 150, 140, 61, 88, 51, 98, 149, 0\end{tabular} \\ \hline
		Second                  & \begin{tabular}[c]{@{}l@{}}0, 152, 177, 88, 131, 55, 90, 62, 198, 13, 33,\\ 44, 98, 97, 112, 9, 118, 129, 99, 14\end{tabular}     \\ \hline
		Third                   & \begin{tabular}[c]{@{}l@{}}103, 85, 24, 124, 41, 11, 47, 194, 74, 31, 64,\\ 49, 18, 75, 8, 54, 12, 181, 80, 117\end{tabular}      \\ \hline
	\end{tabular}
	\label{tab:TinyImageNet}
\end{table}

\begin{table*}[!htb]
	\renewcommand{\arraystretch}{1}
	\centering
	%	\normalsize
	\caption{ The results of adversarial defense (closed-set accuracy) on white-box attacks. }
	\begin{tabular}{c|c|c|c|c|c|c}
		\hline
		\multirow{2}{*}{Method}    & \multicolumn{2}{c|}{\textbf{SVHN}} & \multicolumn{2}{c|}{\textbf{CIFAR-10}} & \multicolumn{2}{c}{\textbf{TinyImageNet}} \\ \cline{2-7} 
		& FGSM         & PGD        & FGSM           & PGD          & FGSM             & PGD            \\ \hline
		Clean                & 96.03$\pm$0.69 &96.03$\pm$0.69  &93.12$\pm$1.83 &93.12$\pm$1.83   &56.80$\pm$3.64 & 56.80$\pm$3.64                \\ 
		Adv on Clean         & 41.61$\pm$3.29 &39.32$\pm$1.82  &31.85$\pm$4.54 &13.02$\pm$4.01   &11.27$\pm$2.61 & 4.41$\pm$0.87                \\ \hline
		Adversarial Training & 88.57$\pm$2.70 &75.82$\pm$2.58  &87.37$\pm$1.15 &72.47$\pm$4.66   &66.61$\pm$1.25 & 40.38$\pm$2.33               \\ 
		Feature Denoising    & 86.94$\pm$3.77 &75.51$\pm$2.68  &87.40$\pm$2.35 &72.55$\pm$4.54   &64.52$\pm$1.36 & 39.35$\pm$3.03               \\ 
		OSDN        	& 89.31$\pm$0.77 &77.97$\pm$1.68  &88.22$\pm$2.97 &74.24$\pm$4.37   &75.17$\pm$7.94 & 41.63$\pm$2.26               \\ \hline
		\textbf{Ours w/ DADL}   & 90.66$\pm$0.18 &80.01$\pm$1.94  &91.91$\pm$4.23 &75.11$\pm$4.91   &76.40$\pm$1.77 & 41.53$\pm$3.12     \\
		\textbf{Ours w/ DADL+CAML}   & \textbf{90.70}$\pm$0.89 &\textbf{81.10}$\pm$1.58  &\textbf{93.40}$\pm$3.44 &\textbf{76.25}$\pm$4.02   &\textbf{81.22}$\pm$4.55 & \textbf{43.77}$\pm$1.57               \\ \hline
	\end{tabular}
	\label{tbl:osacc}
\end{table*}

\begin{table*}[!htbp]
	\renewcommand{\arraystretch}{1}
	\centering
	%	\normalsize
	\caption{ The results of open-set classification (area under the ROC curve) on white-box attacks. }
	\begin{tabular}{c|c|c|c|c|c|c}
		\hline
		\multirow{2}{*}{Method}    & \multicolumn{2}{c|}{\textbf{SVHN}} & \multicolumn{2}{c|}{\textbf{CIFAR-10}} & \multicolumn{2}{c}{\textbf{TinyImageNet}} \\ \cline{2-7} 
		& FGSM         & PGD        & FGSM           & PGD          & FGSM             & PGD            \\ \hline
		Clean                &  91.31$\pm$2.42 & 91.31$\pm$2.42 &  81.20$\pm$2.96   & 81.20$\pm$2.96   & 59.50$\pm$0.89  &59.50$\pm$0.89              \\ 
		Adv on Clean       &  56.44$\pm$1.26 & 54.13$\pm$2.91 &  51.52$\pm$2.81   & 45.56$\pm$0.55   & 47.98$\pm$2.76  &48.60$\pm$1.32            \\\hline
		Adversarial Training &  61.43$\pm$8.08 & 65.25$\pm$4.05 &  75.29$\pm$1.20   & 68.79$\pm$3.23   & 65.10$\pm$8.18   &56.57$\pm$0.96 \\ 
		Feature Denoising    &  64.58$\pm$14.70  &64.92$\pm$4.25&  76.94$\pm$3.70   & 69.83$\pm$2.48   & 65.34$\pm$5.18    &56.12$\pm$1.64     \\ 
		OSDN                 &  71.41$\pm$4.23  & 71.64$\pm$2.67&  79.10$\pm$1.06   & 70.66$\pm$1.79   & 70.81$\pm$5.12 &58.25$\pm$1.90  \\ \hline
		\textbf{Ours w/ DADL}   & 76.77$\pm$1.97  & 73.13$\pm$4.80&  81.41$\pm$1.64   & 73.54$\pm$2.70   & 74.05$\pm$9.51 &58.46$\pm$0.28  \\ 
		\textbf{Ours w/ DADL+CAML}  & \textbf{79.38}$\pm$2.79  & \textbf{74.52}$\pm$2.80&  \textbf{82.06}$\pm$0.86   & \textbf{74.42}$\pm$2.05   & \textbf{76.76}$\pm$5.06 &\textbf{59.94}$\pm$0.75   \\ \hline
	\end{tabular}
	\label{tbl:osauc}
\end{table*}

\begin{table}[!htb]
	\renewcommand{\arraystretch}{1}
	\centering
	%	\normalsize
	\caption{ The results of adversarial defense (closed-set accuracy) on white-box attacks in ImageNet dataset.}
	\begin{tabular}{c|p{2cm}<{\centering}|p{2cm}<{\centering}}
		\hline
		\multirow{2}{*}{Method} & \multicolumn{2}{c}{\textbf{ImageNet}}            \\ \cline{2-3} 
		& FGSM                    & PGD                     \\ \hline
		Clean                   & 64.76          & 64.76         \\
		Adv on Clean            & 4.38          &  0.36         \\ \hline
		Adversarial Training    & 70.22         & 33.26         \\
		Feature Denoising       & 46.48         & 32.40          \\ \hline
		\textbf{Ours} & $\textbf{71.74}$ & $\textbf{38.42}$ \\ \hline
	\end{tabular}
	\label{tbl:osacc_imagenet}
\end{table}

\begin{table}[!htbp]
	\renewcommand{\arraystretch}{1}
	\centering
	%	\normalsize
	\caption{ The results of open-set classification (area under the ROC curve) on white-box attacks in ImageNet dataset. }
	\begin{tabular}{c|p{2cm}<{\centering}|p{2cm}<{\centering}}
		\hline
		\multirow{2}{*}{Method} & \multicolumn{2}{c}{\textbf{ImageNet}} \\ \cline{2-3} 
		& FGSM               & PGD               \\ \hline
		Clean                   & 68.52     & 68.52    \\
		Adv on Clean            & 48.74     & 49.81    \\ \hline
		Adversarial Training    & 60.88     & 54.05    \\
		Feature Denoising       & 63.34     & 52.64    \\ \hline
		\textbf{Ours} & $\textbf{72.10}$     & $\textbf{57.00}$    \\ \hline
	\end{tabular}
	\label{tbl:osauc_imagenet}
\end{table}

\subsection{Datasets}

Following~\cite{yoshihashi2019classification,oza2019c2ae}, the evaluation of our method and other state-of-the-arts are conducted on three standard images classification datasets available for open-set recognition:

\subsubsection{SVHN and CIFAR10} Both CIFAR10~\cite{cifar10Hinton} and SVHN~\cite{netzer2011reading} are classification datasets with 10 classes.  Street-View House Number dataset (SVHN) contains house number signs extracted from Google Street View. CIFAR10 contains images from four vehicle classes and six animal classes. We randomly split 10 classes into 6 known classes and 4 open-set classes to simulate the open-set recognition scenario. We consider three randomly selected splits for testing.

\subsubsection{TinyImageNet} TinyImageNet contains a sub-set of 200 classes selected from ImageNet dataset~\cite{deng2009imagenet}. 20 classes are randomly selected to be known and the remaining 180 classes are chosen to be open-set classes. We consider three randomly chosen splits for evaluation.

The selected known classes and the corresponding splits from SVHN, CIFAR10 and the TinyImageNet datasets are shown in Tables~\ref{tab:svhn}, ~\ref{tab:cifar10}, and \ref{tab:TinyImageNet}, respectively.

\subsubsection{ImageNet} We further carry out experiments on a large-scale ImageNet dataset~\cite{deng2009imagenet}. The ImageNet dataset consists of real-world images with 1000 classes. Following the split proportion between known and open-set classes in TinyImageNet dataset, 100 classes are randomly selected to be known and the remaining 900 classes are chosen to be open-set classes.

\begin{table*}[!htb]
	%	\normalsize
	\renewcommand{\arraystretch}{1}
	\centering
	\caption{Performance of out-of-distribution object detection in CIFAR10 dataset.}
	\begin{tabular}{c|c|c|c|c}  
		\hline
		Detector             & ImageNet-Crop  & ImageNet-Resize & LSUN-Crop      & LSUN-Resize    \\ \hline
		Clean                & 78.9          & 76.2           & 82.1          & 78.7          \\ 
		Adv on Clean         & 4.7          & 4.4           & 7.3          & 3.8          \\ \hline
		Adversarial Training & 35.2          & 34.5           & 35.0           & 34.7          \\ 
		Feature Denoising    & 43.2          & 41.0            & 43.5          & 41.2          \\
		OSDN        & 46.5 		& 44.8  		& 47.1 			& 44.2 \\ \hline
		\textbf{Ours w/ DADL}        &  60.1		& 52.3  		& 60.5			&52.5  \\ 
		\textbf{Ours w/ DADL+CAML}        & \textbf{65.9} 		& \textbf{58.1} 		& \textbf{66.7}	& \textbf{58.7} \\ \hline
	\end{tabular}
	\label{tbl:f1}
\end{table*}
%\noindent \textbf{Implementation Details.} 
\subsection{Implementation Details}
Structure of Resnet-18~\cite{Kaiming_Resnet_CVPR2016} is adopted in our paper, in which encoder network consists of four main blocks. Each main convolutional block in the encoder is embedded with a denoising layer. The decoder network proposed in~\cite{neal2018open} is adopted for the decoder part in our network, which consists of three transpose-convolution layers for experiments in SVHN and CIFAR10 datasets, and four transpose-convolution layers for experiments in TinyImageNet dataset. Both open-set classifier and transformation classifier is composed of a single fully-connected layer. We adopt the standard structure of Resnet-18 as the peer learner.  Adam optimizer~\cite{kingma2014adam} is used and its learning rate is set to $1e-3$. We use the iteration $T= 5$, step size $\epsilon_{step} = 0.01$ for the PGD attacks, and $\epsilon=0.3$ for the FGSM attacks in both adversarial training and testing. We select the trained model for testing based on the best closed-set accuracy on the validation set.

\subsection{Baseline Methods}
We consider the following two recent adversarial defense methods which are most closely related to the proposed method as baselines: \textbf{Adversarial Training}~\cite{madry2017towards} and \textbf{Feature Denoising}~\cite{xie2019feature}. We also compare the performance of our improved method with our initial approach \textbf{OSDN}~\cite{Shao_2020_OSAD}. For the sake of a fair comparison in open-set recognition, an OpenMax layer is integrated on the top of the last hidden layer during testing for all baselines. Moreover, to evaluate the performance of a classifier without a defense mechanism, a Resnet-18 network is trained with clean images collected from known classes and an OpenMax layer is added during testing. In inference, this network is evaluated with clean images, which is denoted as \textbf{clean}. Furthermore, we test this model with adversarial images and this test case is denoted as \textbf{adv on clean}.

\subsection{Protocols}

Conventional open-set recognition requires the model to perform two tasks. First, it should be able to detect open-set samples effectively. Secondly, it should be able to perform correct classification on closed set samples. We thus evaluate the performance of our method and baselines on these two tasks under adversarial attacks. In particular, following previous open-set works~\cite{neal2018open}, we adopt the metric of area under the ROC curve (AUC-ROC) to test the performance on open-set samples detection. On the other hand, to examine the classification ability on closed set samples, we calculate prediction accuracy by only considering known-set samples in the test set. In our experiments, both known and open-set samples are subjected to adversarial attacks prior to testing. White-box attacks, black-box attacks and rectangular occlusion attacks are considered to attack the model. Adversarial samples are generated from known classes using the ground-truth labels, while adversarial samples are generated from open-set classes based on model's prediction.

\subsection{Against White-box Attacks}

\subsubsection{Open-set Recognition}

In experiments regarding white-box attacks, we consider FGSM and PGD for attacking the model. We tabulate the obtained performance for closed-set accuracy and open-set detection in Tables~\ref{tbl:osacc},~\ref{tbl:osauc},~\ref{tbl:osacc_imagenet} and~\ref{tbl:osauc_imagenet}, respectively. Based on the aforementioned analysis, compared to our conference version, the proposed journal version has two major improvements --  Dual-Attentive Denoising Layers (DADL) and Clean-Adversarial Mutual Learning (CAML). We evaluate the proposed method based on these two improvements and denote their experimental results as: Ours w/ DADL and Ours w/ DADL+CAML. From Tables~\ref{tbl:osacc} and~\ref{tbl:osauc}, it can be seen that networks trained on clean images produce very high recognition performance for clean images under both scenarios. However, when the adversarial noise is presented, both open-set detection and closed-set classification performance drops significantly. This validates that current adversarial attacks can easily fool an open-set recognition method such as OpenMax, and thus OSAD is a critical research problem. Three baseline defense mechanisms considered are able to improve the recognition on both known and open-set samples.  It can be observed from Tables~\ref{tbl:osacc} and~\ref{tbl:osauc}, that the proposed method obtains the best open-set detection performance and closed-set accuracy compared to all considered baselines across all three datasets. Specifically, the proposed method with dual-attentive denoising layers (Ours w/ DADL) performs better than our conference version method OSDN and has achieved about $1-5\%$ improvements in adversarial defense and open-set detection across all datasets. This demonstrates that the proposed dual-attentive denoising layers are more advanced and effective to address the research problem of OSAD compared to the non-local filters based denoising layers proposed in our conference version. In addition, as shown in the results of Ours w/ DADL+CAML from Tables~\ref{tbl:osacc} and~\ref{tbl:osauc}, the performance can be further improved after incorporating the peer learner to carry out clean-adversarial mutual learning (Ours w/ DADL+CAML). In particular, the proposed method with both improvements (Ours w/ DADL+CAML) can further improve the performance by $5\%$ in FGSM adversarial defense in TinyImageNet dataset and about $3\%$ in open-set detection under FGSM attacks in SVHN dataset compared to the proposed method only with Dual-Attentive Denoising Layers (Ours w/ DADL). In other datasets, this improvement varies between $1-2\%$. This shows the effectiveness of the proposed Clean-Adversarial Mutual Learning and thus the knowledge exploited from the peer learner about clean image classification can benefit both adversarial defense and open-set recognition.
 
%It is interesting to note that methods involving adversarial training perform better than the baseline of clean image classification under FGSM attacks on the TinyImageNet dataset. This is because only 20 classes from the TinyImageNet dataset are selected for training and each class has only 500 images.  When a small dataset is used to train a model with large number of parameters, it is easier for the network to overfit to the training set. Such network observes variety of data in the presence of adversarial training.  Therefore model reaches a more generalizable optimization solution during training. 

Moreover, as shown in Tables~\ref{tbl:osacc_imagenet} and~\ref{tbl:osauc_imagenet}, in the more challenging large-scale ImageNet dataset, the proposed method also performs better than the other baselines in terms of both close-set recognition and open-set detection. This further demonstrates the proposed method is also able to address the OSAD problem in real-world scenarios.

\subsubsection{Out-of-distribution detection}

This section evaluates the performance of the proposed method on the out-of-distribution detection (OOD)~\cite{hendrycks2016baseline} task in CIFAR10, in which the protocol described in~\cite{yoshihashi2019classification} is used. All classes in CIFAR10 are regarded as known classes and test images from ImageNet and LSUN datasets~\cite{yu2015lsun} (both cropped and resized images) are treated as out-of-distribution classes~\cite{liang2017enhancing}. The performance of adversarial defense is tested with the adversarial images produced by the PGD attacks for both known and OOD data. Adversarial samples from the known classes are generated based on the ground-truth labels, while adversarial samples from the OOD class are generated using the model's prediction. Macro-averaged F1 score is adopted as the evaluation metric. OpenMax layer with a threshold $0.95$ is used when assigning open-set labels to the query images. We tabulate the OOD detection performance in Table~\ref{tbl:f1}, in which all three baselines and the proposed method are compared across all four cases. From Table~\ref{tbl:f1}, it is evident that the proposed method outperforms other baselines across all the test cases regarding the ODD task. In particular, the improvements vary between $14-20\%$ across all ODD datasets, which are significant improvements compared to other baselines. This experiment further demonstrates the effectiveness of our method for the open-set samples detection in the presence of adversarial attacks.

\begin{table}[!htb]
	%	\normalsize
	\renewcommand{\arraystretch}{1}
	\centering
	\caption{Results corresponding to the ablation study. }
	\begin{tabular}{p{0.2cm}<{\centering}p{0.2cm}<{\centering}c<{\centering}p{0.2cm}<{\centering}c|c|c}
		\hline
		\multicolumn{5}{c|}{\multirow{2}{*}{\textbf{Methods}}} & \multicolumn{2}{c}{\textbf{CIFAR-10}}          \\ \cline{6-7} 
		\multicolumn{5}{c|}{}                                  & \textbf{AUC-ROC} & \textbf{Close-set Acc} \\ \hline
		\multicolumn{5}{c|}{Clean}                             & 83.72            & 92.79                        \\ \hline
		\multicolumn{5}{c|}{Adv on Clean}                      & 45.98            & 8.65                         \\ \hline
		\multicolumn{5}{c|}{Components of Ours}                & \textbf{}        &                              \\ 
		Enc       & Dec       & DADL       & SSD       & CAML      & \textbf{}        & \textbf{}                    \\ \hline
		\checkmark         &           &           &           &         & 66.10            & 69.90                        \\
		\checkmark         &           & \checkmark         &           &         &    69.86              & 70.00                             \\ 
		\checkmark         & \checkmark         &           &           &         & 67.34            & 68.85                        \\ 
		\checkmark         & \checkmark         & \checkmark         &           &         &    70.52              &  70.98                            \\ 
		\checkmark         &           & \checkmark         & \checkmark         &         & 70.00            & 72.80                        \\ 
		\checkmark         & \checkmark         &           & \checkmark         &         & 67.04            & 72.10                        \\ 
		\checkmark         & \checkmark         &           & \checkmark         &\checkmark          & 69.87        & 72.20                     \\
		\checkmark         & \checkmark         & \checkmark         & \checkmark         &         & 71.69            & 73.13                        \\ 
		\checkmark         & \checkmark         & \checkmark         & \checkmark         & \checkmark       & \textbf{73.72}            & \textbf{74.14}                        \\ \hline
	\end{tabular}
	\label{tabAblationStudy}
\end{table}

\begin{table}[!htb]
	\renewcommand{\arraystretch}{1}
	\centering
	%	\normalsize
	\caption{ Results of adversarial defense (closed-set accuracy) compared to method with inputs mixed up with clean and adversarial images. }
	\begin{tabular}{c|c|c}
		\hline
		\multirow{2}{*}{Method} & \multicolumn{2}{c}{\textbf{CIFAR-10}}            \\ \cline{2-3} 
		& FGSM                    & PGD                     \\ \hline
		Clean                   & 93.12$\pm$1.83 			&93.12$\pm$1.83           \\
		Adv on Clean            &31.85$\pm$4.54 			&13.02$\pm$4.01          \\ \hline
		Clean Adv Mixup    		&61.44$\pm$0.89         	&39.24$\pm$4.37          \\
		\textbf{Ours} & \textbf{93.40}$\pm$3.44 &\textbf{76.25}$\pm$4.02 \\ \hline
	\end{tabular}
	\label{tabMixupAcc}
\end{table}

\begin{table}[!htbp]
	\renewcommand{\arraystretch}{1}
	\centering
	%	\normalsize
	\caption{  Results of open-set classification (area under the ROC curve) compared to method with inputs mixed up with clean and adversarial images. }
	\begin{tabular}{c|c|c}
		\hline
		\multirow{2}{*}{Method} & \multicolumn{2}{c}{\textbf{CIFAR-10}} \\ \cline{2-3} 
		& FGSM               & PGD               \\ \hline
		Clean                   &  81.20$\pm$2.96   & 81.20$\pm$2.96   \\
		Adv on Clean            &  51.52$\pm$2.81   & 45.56$\pm$0.55    \\ \hline
		Clean Adv Mixup       & 62.90$\pm$2.26     & 56.10$\pm$3.73    \\
		\textbf{Ours} & \textbf{82.06}$\pm$0.86   & \textbf{74.42}$\pm$2.05    \\ \hline
	\end{tabular}
	\label{tabMixupAUROC}
\end{table}

\subsubsection{Ablation Study}
The proposed network consists of five CNN components with four branches. Specifically, we denote the feature encoding carried out in encoder as Enc for short, the feature denoising carried out by dual-attentive denoising layers as DADL for short, image generation carried out in the decoder as Dec for short, self-supervised denoising carried out by the transformation classifier as SSD for short, and clean-adversarial mutual learning carried out by the peer learner as CAML for short. In this section, we investigate the impact of each network component to the overall performance of the system. To validate the effectiveness of various parts integrated in our proposed network, this section conducts the ablation study of our network using CIFAR10 dataset for the task of open-set recognition under PGD attacks. Considered cases and the corresponding results obtained for each case are tabulated Table~\ref{tabAblationStudy}. From Table~\ref{tabAblationStudy}, it can be seen that all the aforementioned four branches are necessary and effective for the improvement of both adversarial defense and open-set recognition. In particular, Table~\ref{tabAblationStudy} shows that the proposed dual-attentive denoising layers are critical and the performance can be significantly improved with it. Moreover, compared to the network without CAML (6-th row of Table~\ref{tabAblationStudy}), adding this component (7-th row of Table~\ref{tabAblationStudy}) can achieve better performance in terms of both close-set accuracy and open-set detection. This further demonstrates the effectiveness of CAML. Furthermore, from the last row of Table~\ref{tabAblationStudy}, it can be seen that the overall performance can be further improved with the CAML, which clearly demonstrates the contribution of this component to the whole network. In all, Table~\ref{tabAblationStudy} shows that added components complement each other to produce better performance for both adversarial defense and open-set recognition.

\subsubsection{Method with inputs mixed up with clean and adversarial images}
Compared to our conference version which only exploits adversarial images to solve OSAD, one of the key improvements of the proposed method is to explore more complementary knowledge from clean image classification in the peer learner via clean-adversarial mutual learning to aid adversarial defense and open-set recognition. To  demonstrate the effectiveness of the clean-adversarial mutual learning, this section compares one of the other possible ways to exploit the information of clean image classification. Specifically, regarding this comparison method, we remove the peer learner branch and input  clean images directly into the encoder for classification. All the other parts of the proposed method remain the same. In this case, the inputs of this comparison method are mixed up with adversarial and corresponding clean images. Results corresponding to this experiment using CIFAR10 dataset, denoted as Clean Adv Mixup, are shown in Table~\ref{tabMixupAcc} and Table~\ref{tabMixupAUROC}. From Table~\ref{tabMixupAcc} and Table~\ref{tabMixupAUROC}, it can be seen that although this comparison performs better than the method without any adversarial defense (adv on clean), it has much lower performance compared to the proposed method. We argue this is because the encoder embedded with denoising layers has a hard time simultaneously performing feature denoising for adversarial images and feature encoding for the clean images. In this case, the adversarial training based on adversarial images and normal supervised training based on clean images are hard to complement each other and may even contradict with each other. Comparatively, the proposed clean-adversarial mutual learning lets the two independent branches, which carry out adversarial training and supervised training respectively, mutually learn with each other. Therefore, more complementary knowledge about the clean image classification is easier to exploit from the peer learner to accurately aid adversarial defense and open-set recognition.

\subsection{Against Black-box Attacks}

\begin{table}[!htb]
	\renewcommand{\arraystretch}{1}
	\centering
	%	\normalsize
	\caption{  The results of adversarial defense (closed-set accuracy) on black-box attacks in SVHN dataset.}
	\begin{tabular}{c|c|c}
		\hline
		\multirow{2}{*}{Method} & \multicolumn{2}{c}{\textbf{SVHN}}            \\ \cline{2-3} 
		& FGSM                    & PGD                     \\ \hline
		Clean                   & 96.03$\pm$0.70          & 96.03$\pm$0.70          \\
		Adv on Clean            & 28.29$\pm$13.03          & 29.37$\pm$13.93          \\ \hline
		Adversarial Training    & 80.20$\pm$4.15         & 80.05$\pm$3.02          \\
		Feature Denoising       & 80.51$\pm$3.92          & 80.64$\pm$2.68          \\
		OSDN                    & 83.23$\pm$2.78          & 80.94$\pm$1.61          \\ \hline
		\textbf{Ours w/ DADL}    & 82.67$\pm$1.22          & 81.75$\pm$2.06          \\
		\textbf{Ours w/ DADL+CAML} & \textbf{85.90}$\pm$2.46 & \textbf{82.19}$\pm$2.12 \\ \hline
	\end{tabular}
	\label{tabBlackSVHNAcc}
\end{table}

\begin{table}[!htbp]
	\renewcommand{\arraystretch}{1}
	\centering
	%	\normalsize
	\caption{  The results of open-set classification (area under the ROC curve) on black-box attacks in SVHN dataset. }
	\begin{tabular}{c|c|c}
		\hline
		\multirow{2}{*}{Method} & \multicolumn{2}{c}{\textbf{SVHN}} \\ \cline{2-3} 
		& FGSM               & PGD               \\ \hline
		Clean                   & 91.31$\pm$2.42     & 91.31$\pm$2.42    \\
		Adv on Clean            & 52.79$\pm$3.57     & 53.09$\pm$3.53    \\ \hline
		Adversarial Training    & 76.66$\pm$2.15     & 69.48$\pm$3.21    \\
		Feature Denoising       & 77.49$\pm$4.07     & 69.69$\pm$3.61    \\
		OSDN                    & 79.99$\pm$2.40     & 76.46$\pm$1.36    \\ \hline
		\textbf{Ours w/ DADL}    & 78.68$\pm$3.07     & 77.98$\pm$2.01    \\
		\textbf{Ours w/ DADL+CAML} & \textbf{80.27}$\pm$1.81     & \textbf{79.00}$\pm$2.62    \\ \hline
	\end{tabular}
	\label{tabBlackSVHNAUROC}
\end{table}

\begin{table}[!htb]
	\renewcommand{\arraystretch}{1}
	\centering
	%	\normalsize
	\caption{ The results of adversarial defense (closed-set accuracy) on black-box attacks in CIFAR10 dataset. }
	\begin{tabular}{c|c|c}
		\hline
		\multirow{2}{*}{Method} & \multicolumn{2}{c}{\textbf{CIFAR-10}}            \\ \cline{2-3} 
		& FGSM                    & PGD                     \\ \hline
		Clean                   & 93.29$\pm$2.56          & 93.29$\pm$2.56          \\
		Adv on Clean            & 35.68$\pm$8.16          & 36.33$\pm$7.32          \\ \hline
		Adversarial Training    & 79.65$\pm$10.99         & 81.98$\pm$5.76          \\
		Feature Denoising       & 80.06$\pm$7.68          & 82.78$\pm$4.41          \\
		OSDN                    & 82.20$\pm$6.50          & 84.27$\pm$4.44          \\ \hline
		\textbf{Ours w/ DADL}    & 81.29$\pm$4.99          & 84.55$\pm$4.52          \\
		\textbf{Ours w/ DADL+CAML} & \textbf{83.01}$\pm$6.92 & \textbf{84.93}$\pm$4.87 \\ \hline
	\end{tabular}
	\label{tabBlackCifar10Acc}
\end{table}

\begin{table}[!htbp]
	\renewcommand{\arraystretch}{1}
	\centering
	%	\normalsize
	\caption{ The results of open-set classification (area under the ROC curve) on black-box attacks in CIFAR10 dataset. }
	\begin{tabular}{c|c|c}
		\hline
		\multirow{2}{*}{Method} & \multicolumn{2}{c}{\textbf{CIFAR-10}} \\ \cline{2-3} 
		& FGSM               & PGD               \\ \hline
		Clean                   & 79.95$\pm$2.84     & 79.95$\pm$2.84    \\
		Adv on Clean            & 51.28$\pm$1.22     & 52.82$\pm$1.81    \\ \hline
		Adversarial Training    & 70.49$\pm$9.70     & 72.56$\pm$1.01    \\
		Feature Denoising       & 71.63$\pm$3.95     & 75.37$\pm$3.97    \\
		OSDN                    & 73.61$\pm$0.98     & 76.68$\pm$5.68    \\ \hline
		\textbf{Ours w/ DADL}    & \textbf{75.35}$\pm$3.01     & 76.62$\pm$2.72    \\
		\textbf{Ours w/ DADL+CAML} & 74.87$\pm$1.93     & \textbf{76.75}$\pm$3.57    \\ \hline
	\end{tabular}
	\label{tabBlackCifar10AUROC}
\end{table}

Several works have demonstrated a phenomenon that adversarial examples generated for one model can be  misclassified by the other models. This phenomenon is known as transferability~\cite{liu2016delving,papernot2017practical} which can be leveraged to design the evaluation corresponding to black-box attacks~\cite{liu2016delving}. Specifically, to attack a target model trained on one dataset, we first train a substitute model in a different network structure using adversarial training with white-box attacks on the same training data. Then we perform the white-box attacks to the substitute model on the test data to generate adversarial examples. Finally, we measure the accuracy of the target model on these generated adversarial test data to evaluate their performance against black-box attacks.

We use Resnet-34 and VGG-13 as the substitute models in the black-box experiments using SVHN and CIFAR10 datasets to generate adversarial samples, respectively. Adversarial examples from the substitute models are generated by FGSM and PGD attacks. We use the three baselines and the proposed network as the target models to evaluate their black-box defense performance on these generated adversarial samples. We tabulate the performance for closed-set accuracy and open-set detection under black-box attacks in Tables~\ref{tabBlackSVHNAcc}, ~\ref{tabBlackSVHNAUROC}, ~\ref{tabBlackCifar10Acc}, and~\ref{tabBlackCifar10AUROC}. From these Tables, it can be seen that the proposed method obtains the best performance of both adversarial defense and open-set recognition across all considered cases under the black-box attack scenario. In particular, the proposed method can achieve about $1-2\%$ improvements in the black-box experiment in SVHN dataset.  This experiment shows that the proposed method is more robust against various black-box attacks and is able to generalize well to open-set samples under black-box attacks.

\subsection{Against Rectangular Occlusion Attack}

\begin{table}[!htb]
	\renewcommand{\arraystretch}{1}
	\centering
	%	\normalsize
	\caption{ The results of adversarial defense (closed-set accuracy) and open-set classification (area under the ROC curve) on rectangular occlusion attack in CIFAR10 dataset. }
	\begin{tabular}{c|c|c}
		\hline
		\multirow{2}{*}{Method} & \multicolumn{2}{c}{\textbf{CIFAR-10}}            \\ \cline{2-3} 
		& Closed-set Acc                   & AUC-ROC                      \\ \hline
		Clean                   & 93.12$\pm$1.83          & 81.20$\pm$2.96          \\
		Adv on Clean            & 32.93$\pm$3.53         & 54.08$\pm$2.02          \\ \hline
		Adversarial Training    & 87.48$\pm$2.98         & 76.35$\pm$1.87          \\
		Feature Denoising       & 87.55$\pm$2.89          & 77.90$\pm$1.30          \\
		OSDN                    & 88.99$\pm$2.54          & 79.53$\pm$1.73          \\ \hline
		\textbf{Ours w/ DADL}    & 88.62$\pm$2.60          & 78.78$\pm$2.18          \\
		\textbf{Ours w/ DADL+CAML} & \textbf{89.61}$\pm$2.42 & \textbf{79.86}$\pm$1.15 \\ \hline
	\end{tabular}
	\label{tabPatchAttackCifar10}
\end{table}

\begin{table}[!htb]
	\renewcommand{\arraystretch}{1}
	\centering
	%	\normalsize
	\caption{ The results of adversarial defense (closed-set accuracy) and open-set classification (area under the ROC curve) on rectangular occlusion attack in TinyImageNet dataset. }
	\begin{tabular}{c|c|c}
		\hline
		\multirow{2}{*}{Method} & \multicolumn{2}{c}{\textbf{TinyImageNet}}            \\ \cline{2-3} 
		& Closed-set Acc                   & AUC-ROC                     \\ \hline
		Clean                   & 56.80$\pm$3.64          & 59.50$\pm$0.89          \\
		Adv on Clean            & 12.79$\pm$3.92         & 50.30$\pm$1.51          \\ \hline
		Adversarial Training    & 60.25$\pm$1.06         & 64.46$\pm$3.16          \\
		Feature Denoising       & 57.42$\pm$1.41          & 64.30$\pm$0.84          \\
		OSDN                    & 58.98$\pm$1.54          & 65.63$\pm$1.23         \\ \hline
		\textbf{Ours w/ DADL}    & 60.97$\pm$1.19          & 65.78$\pm$1.21          \\
		\textbf{Ours w/ DADL+CAML} & \textbf{63.77}$\pm$1.57 & \textbf{67.39}$\pm$0.45 \\ \hline
	\end{tabular}
	\label{tabPatchAttackTinyImageNet}
\end{table}

To develop the robustness of deep neural networks against physically realizable attacks, such as eyeglass frame attack~\cite{sharif2016accessorize} and sticker attack on stop signs~\cite{eykholt2018robust}, Wu \textit{et al.}~\cite{wu2019defending} propose a new abstract adversarial model, rectangular occlusion attack, where an adversary introduces a small adversarially crafted rectangle in an image. This attack can capture the key common element of various physically realizable attacks that involves the introduction of adversarial occlusions to a part of the input. In this section, we conduct experiments using rectangular occlusion attacks (i.e. physically realizable attacks).  In Tables~\ref{tabPatchAttackCifar10}, ~\ref{tabPatchAttackTinyImageNet} we tabulate both open-set detection performance and closed-set classification accuracy corresponding to rectangular occlusion attacks in CIFAR10 and  TinyImageNet datasets. From Tables~\ref{tabPatchAttackCifar10}, ~\ref{tabPatchAttackTinyImageNet}, it can be seen that the proposed method obtains the best performance on both adversarial defense and open-set recognition under rectangular occlusion attacks. In particular, the proposed method can achieve about $3\%$ and $2\%$ improvements in terms of adversarial defense and open-set recognition, respectively in TinyImageNet dataset compared to the baselines. This experiment clearly shows the great potential of the proposed method against physically realizable attacks.

\subsection{Qualitative Results}
To provide a better understanding on the proposed algorithm, clean, adversarial and reconstructed images are first visualized. We also visualize the obtained latent features in 2D with tSNE visualization. Finally, randomly selected feature maps from the encoder network are displayed to demonstrate the results of feature denoising.

\subsubsection{Images Visualization}
\begin{figure}
	
	\begin{center}
		
		\includegraphics[height=8.2cm, width=0.9\linewidth]{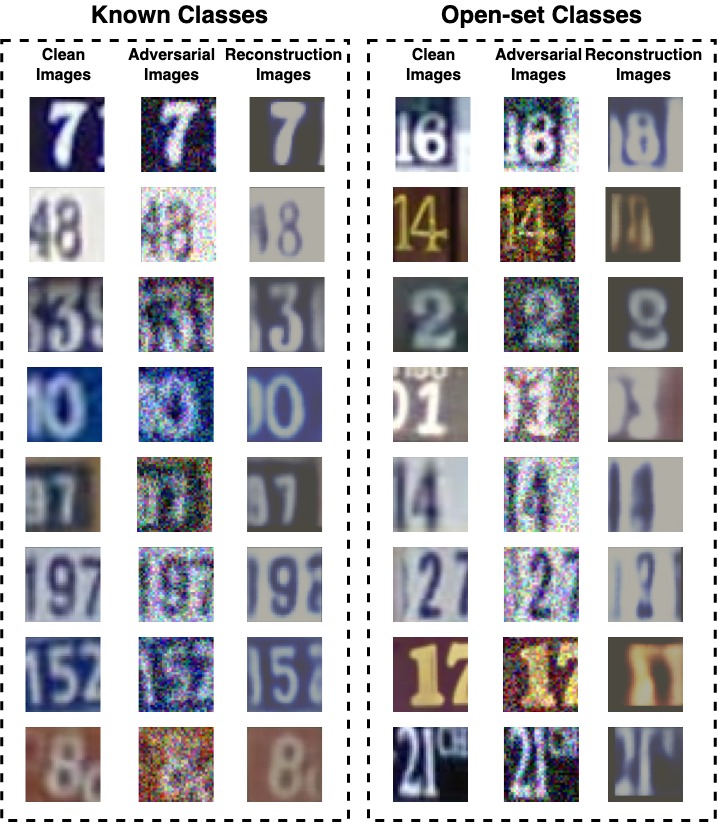}
		
	\end{center}
	\vskip-10pt	\caption{{Visualization of input clean images, corresponding adversarial images, and the reconstructed images from the decoder.}}
	\label{fig:vis_rec}
\end{figure}

\begin{figure*}[!htb] 
	
	\begin{center}
		
		\includegraphics[height=3.7cm, width=0.92\linewidth]{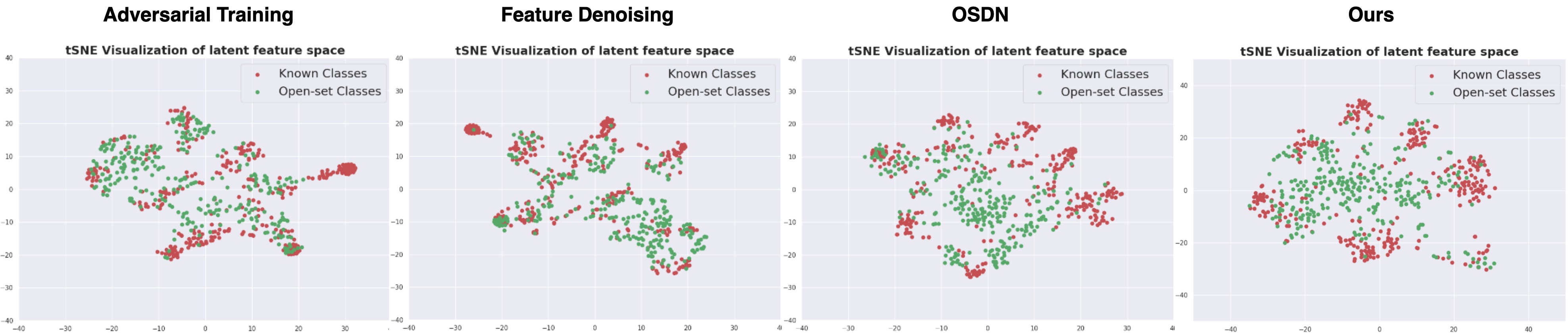}
		
	\end{center}
	\vskip-12pt	\caption{{tSNE visualization of the latent feature space under PGD attack corresponding to our method and baselines.}}
	\label{fig:tsnePGD}
\end{figure*}

\begin{figure*}[!htb] 
	
	\begin{center}
		
		\includegraphics[height=3.7cm, width=0.92\linewidth]{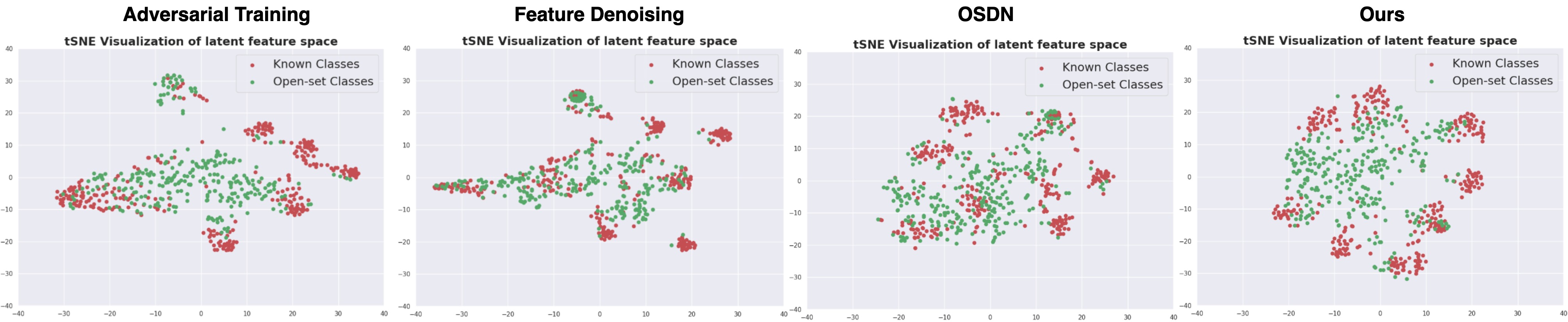}
		
	\end{center}
	\vskip-12pt	\caption{{tSNE visualization of the latent feature space under FGSM attack corresponding to our method and baselines.}}
	\label{fig:tsneFGSM}
\end{figure*}

Visualization is considered in SVHN dataset. In Figure~\ref{fig:vis_rec}, we present a set of clean images, corresponding adversarial images perturbed by the PGD attacks and images decoded from the latent features with our proposed network. Visualizations regarding known and open-set samples are presented in two columns, respectively. By comparing the samples between columns of adversarial images and reconstruction images in Figure~\ref{fig:vis_rec}, it can be observed that image noise has been clearly reduced in both open-set and known-class images. However, the reconstruction quality is inferior for the open-set samples compared to the known class samples. Reconstructions of open-set samples look blurry and structurally abnormal. For example, in the block of Open-set Classes, the image of digit 2 shown in the third row, looks like an intermediate shape between digit 3 and 8 once reconstructed. We will further explain the reason for this phenomenon in the following section.

\subsubsection{Visualization of Latent Features}

Figures~\ref{fig:tsnePGD} and~\ref{fig:tsneFGSM} visualize the latent features generated by the proposed method along with three other baselines using tSNE visualization~\cite{maaten2008visualizing}. The visualizations are carried out considering both FGSM and PGD attacks in SVHN dataset. As shown in Figures~\ref{fig:tsnePGD} and~\ref{fig:tsneFGSM}, compared to all baseline methods, the proposed method enables most of open-set features to lie away from manifold of known set features and achieves less overlap between the two types of features. This visually verifies the better performance regarding open-set detection obtained by the proposed method. In particular, from Figure~\ref{fig:tsnePGD}, it can be observed that our conference version method OSDN has less overlap between the two types of features compared to the baselines of adversarial training and feature denoising under PGD attacks, which is similar to the proposed method. However, in Figure~\ref{fig:tsneFGSM}, the overlap of features corresponding to OSDN increases in the presence of FGSM attacks. In contrast, the proposed method maintains superior separation among features from known samples and open-set samples. This further demonstrates the superiority of the proposed method for open-set recognition under various adversarial attacks compared to our conference version method.

It should be noted that the quality of reconstruction of open-set samples obtained by the decoder network tends to be poor when the corresponding open-set latent features lie away from the manifold of known classes. Therefore, the tSNE plots just justify why the quality of reconstructions is poorer for open-set samples based on our method shown in Figure~\ref{fig:vis_rec}. As such, Figure~\ref{fig:vis_rec} along with Figures~\ref{fig:tsnePGD} and~\ref{fig:tsneFGSM} together show the effectiveness of our method for identifying open-set samples under adversarial attacks.

\subsubsection{Visualization of Features Maps}
\begin{figure*}[!htb] 
	\begin{center}
		\includegraphics[height=8.6cm, width=0.8\linewidth]{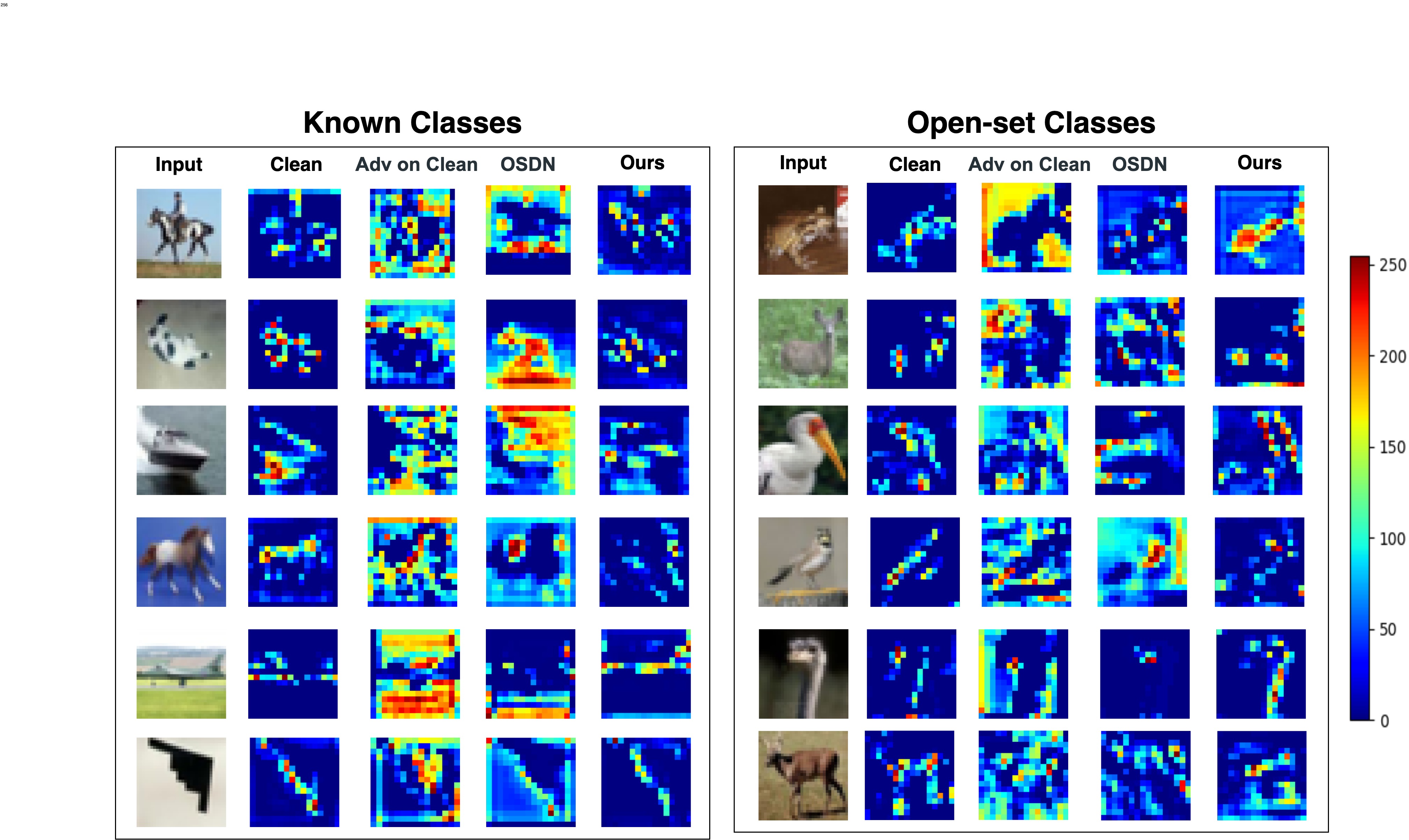}
	\end{center}
	\vskip-10pt	\caption{{Feature map visualization in the res$_2$ block of Resnet-18, encoder of OSDN, and  encoder of proposed network. (Best viewed in colors.)}}
	\label{fig:featmap}
\end{figure*}
To demonstrate the effectiveness of feature denoising carried out in the encoder of the proposed network,  we visualize some randomly selected feature maps of the second residual block from the trained Resnet-18~\cite{Kaiming_Resnet_CVPR2016}, the encoder of our conference version OSDN network and the encoder of the proposed network. The feature maps for each row of examples are from the same corresponding channel of the feature maps. We consider samples from both known and open-set classes from CIFAR10 dataset. Figure~\ref{fig:featmap} shows a set of feature maps of the trained Resnet-18 applied on the clean images (denoted as \textbf{Clean}) and the corresponding PGD adversarial images (denoted as \textbf{Adv on Clean}). From samples of Resnet-18 in Figure~\ref{fig:featmap}, it can be observed that feature maps of clean images mainly focus on semantically informative regions, while feature maps corresponding to the adversarial images have noisy activations on semantically irrelevant regions. This quantitatively demonstrates that a lot of adversarial noise is produced in the features as the adversarial images are propagated through the network~\cite{liao2018defense}.  Figure~\ref{fig:featmap} further shows the feature maps corresponding to the proposed method applied on the same PGD adversarial images (denoted as \textbf{Ours}). From samples of the proposed method in Figure~\ref{fig:featmap}, it can be observed that compared to Resnet-18 without any defense mechanism, the proposed network is able to reduce adversarial noise significantly in feature maps of adversarial images. In particular, as illustrated in Figure~\ref{fig:featmap}, compared to our conference version OSDN, the resulting denoised feature maps of the proposed method are closer to the feature maps corresponding to the clean images. For example, from the third row of known classes block and fourth row of open-set classes block, it can be observed that the resulting feature maps corresponding to OSDN are still filled with adversarial noise. Comparatively, the proposed method is able to remove the noise and restore the feature maps. Moreover, from the fifth row of known classes block and the fifth row of open-set classes block, it can be observed that the semantic regions of feature maps are lost after feature denoising with OSDN. In contrast, the proposed method can maintain the semantic regions while removing the adversarial noise. This means that the proposed method carries out feature denoising more accurately. This visualization further demonstrates that the proposed network indeed carries out the feature denoising through the embedded feature denoising layers, and obtains much better adversarial robustness. In addition, the visualization can also verify that the proposed dual-attentive denoising layer in the proposed method can achieve better feature denoising performance compared to the non-local means-based filter adopted in our conference version.

\section{Conclusion}
A novel research problem -- Open-set Adversarial Defense (OSAD) is studied in this paper. In this new problem, we observe existing adversarial defense mechanisms fail to generalize well in the presence of open-set samples. On the other hand, open-set classifiers are vulnerable to existing adversarial attack mechanisms. We propose an Open-Set Defense Network with Clean-Adversarial Mutual Learning (OSDN-CAML) which aims to identify the open-set samples in the presence of adversarial attacks. The proposed network consists of a feature denoising operation with dual-attentive denoising layers, a self-supervision function, a clean image generation operation and a clean-adversarial mutual learning mechanism. Extensive experiments carried out on four publicly available classification datasets validate the superiority of the proposed method to deal with both open-set detection and various adversarial defenses. In addition, the task of out-of-distribution detection under the adversarial setting is also shown to be well addressed by the proposed method. Finally, various visualizations corresponding to different properties of the proposed method are presented, which provide a more comprehensive understanding of the proposed method. In the future, we plan to further extend the current OSDN-CAML to defend more types of adversarial attacks including adaptive adversarial attacks for the OSAD problem.

\begin{acknowledgements}
This work is partially supported by Research Grants Council (RGC/HKBU12200820), Hong Kong. Vishal M. Patel was supported by an ARO grant W911NF-21-1-0135.
\end{acknowledgements}


\bibliographystyle{spmpsci}      % mathematics and physical sciences
\bibliography{ref}   % name your BibTeX data base

%\begin{thebibliography}{}
%%
%% and use \bibitem to create references. Consult the Instructions
%% for authors for reference list style.
%\end{thebibliography}

\end{document}